\documentclass[conference]{IEEEtran}

\usepackage{graphicx}
\usepackage{amsmath}
\usepackage{amssymb}    
\usepackage{pifont}     
\usepackage{array}      
\usepackage{booktabs}   
\usepackage{comment}
\usepackage{footmisc}
\usepackage{float}
\usepackage{xurl} 
\usepackage{hyperref}

\IEEEoverridecommandlockouts
\usepackage{enumitem} 

\usepackage{array}
\usepackage{tabularx}
\usepackage{caption}

\newcolumntype{C}[1]{>{\centering\arraybackslash}p{#1}}

\usepackage{hyperref}

\usepackage{subcaption}

\usepackage{xcolor}
\usepackage{color, soul}

\usepackage{algorithm}
\usepackage{algpseudocode}

\usepackage{mdframed}

\newmdenv[
  backgroundcolor=gray!20,
  linecolor=gray!20,
  linewidth=0.25pt,
  roundcorner=10pt,
  skipabove=5pt,
  skipbelow=5pt,
  innertopmargin=2pt,
  innerbottommargin=2pt,
  innerleftmargin=5pt,
  innerrightmargin=5pt,
]{custombox}

\title{The Future of AI \\ Exploring the Potential of Large Concept Models}

\author{
\IEEEauthorblockN{Hussain Ahmad\textsuperscript{†*} \thanks{\textsuperscript{$\dagger$}Corresponding author.}\thanks{*Authors contributed equally to this work.}}
\IEEEauthorblockA{\textit{The University of Adelaide, Australia}\\
hussain.ahmad@adelaide.edu.au}
\and
\IEEEauthorblockN{Diksha Goel\textsuperscript{*}}
\IEEEauthorblockA{\textit{CSIRO's Data61, Australia}\\
diksha.goel@data61.csiro.au}
}

\begin{document}

\maketitle

\begin{abstract}

The field of Artificial Intelligence (AI) continues to drive transformative innovations, with significant progress in conversational interfaces, autonomous vehicles, and intelligent content creation. Since the launch of ChatGPT in late 2022, the rise of Generative AI has marked a pivotal era, with the term \textit{Large Language Models (LLMs)} becoming a ubiquitous part of daily life. LLMs have demonstrated exceptional capabilities in tasks such as text summarization, code generation, and creative writing. However, these models are inherently limited by their token-level processing, which restricts their ability to perform abstract reasoning, conceptual understanding, and efficient generation of long-form content. To address these limitations, Meta has introduced \textit{Large Concept Models (LCMs)}, representing a significant shift from traditional token-based frameworks. LCMs use \textit{concepts} as foundational units of understanding, enabling more sophisticated semantic reasoning and context-aware decision-making. Given the limited academic research on this emerging technology, our study aims to bridge the knowledge gap by collecting, analyzing, and synthesizing existing grey literature to provide a comprehensive understanding of LCMs. Specifically, we (i) identify and describe the features that distinguish LCMs from LLMs, (ii) explore potential applications of LCMs across multiple domains, and (iii) propose future research directions and practical strategies to advance LCM development and adoption.

\end{abstract}


\section{Introduction}

Large Language Models (LLMs) have reshaped the landscape of Artificial Intelligence (AI), emerging as indispensable tools for tasks such as natural language processing, content generation, and complex decision-making \cite{zheng2025towards, chang2024survey}. The launch of ChatGPT in late 2022 was a defining moment, ushering in a new era of Generative AI and integrating LLMs into everyday applications \cite{haque2022think, Mehul_LCM}. At the core of these models is the Transformer architecture, a sophisticated neural network that processes and interprets user prompts \cite{huang2023advancing}. A critical yet often overlooked component in this process is the tokenizer. This mechanism segments input text into smaller units called tokens, which can be words, subwords, or characters mapped to the model’s vocabulary \cite{borgeaud2022improving}. This tokenization step is critical for effective interpretation of context, enabling the Transformer to generate coherent responses \cite{chopra2024chatnvd}. The synergy between the tokenizer and the Transformer architecture underpins the remarkable performance of LLMs, solidifying their position at the forefront of modern AI advancements \cite{gereti2024token}.

Despite these achievements, LLMs face inherent limitations tied to their token-level processing, where predictions are generated one token at a time based on preceding sequences \cite{kim2025palm, Lance_lcm}. This approach constrains their ability to tackle tasks that demand deep reasoning, extended context management, or highly structured outputs \cite{LLM_Limit}. Unlike human cognition, which typically begins with a high-level outline and progressively adds detail, LLMs rely on vast amounts of training data without explicit mechanisms for hierarchical structuring \cite{cuskley2024limitations}. As a result, they often struggle to maintain coherence in long-form content that spans multiple sections \cite{LLM_Limit22}. In addition, the quadratic computational complexity of processing long sequences poses scalability challenges, limiting their efficiency \cite{LLM_Limit333}. While techniques such as sparse attention \cite{fu2024moa} and locality-sensitive hashing \cite{petrick2022locality} have been introduced to address these issues, they provide partial solutions and do not fully resolve the underlying constraints. Therefore, advancing LLMs requires novel approaches that integrate explicit hierarchical reasoning for well-structured, contextually consistent outputs.

To overcome the limitations of traditional LLMs, Meta\footnote{\url{https://about.meta.com}} has introduced\textbf{\textit{ Large Concept Models}} (LCMs)\footnote{\url{https://github.com/facebookresearch/large_concept_model}} \cite{the2024large}, a groundbreaking framework that shifts the fundamental unit of processing from individual tokens to entire semantic units, referred to as \emph{\textbf{concepts}} \cite{video_lcm}. Unlike LLMs, which predict words or subwords sequentially \cite{LLMsKnow}, LCMs operate at a higher level of abstraction, representing and reasoning about complete ideas \cite{Richard_lcm}. By grouping sentences or conceptual clusters, LCMs can more efficiently handle long-context tasks and produce outputs that are both coherent and interpretable \cite{LCMsss}. This conceptual approach not only mirrors the way humans organize and process information but also significantly reduces the computational costs associated with managing long sequences \cite{lcm2024}. LCMs can demonstrate exceptional performance in cross-lingual tasks, seamlessly generating and processing text across multiple languages without retraining, and excel in multimodal tasks, integrating text and speech for real-time translation and transcription \cite{pruseth_lcm_agi}. Their ability to synthesize and expand lengthy content with relevant context makes them especially effective in tasks involving extended document comprehension \cite{ashish_lcm}. By shifting focus from token-level to concept-level modelling, LCMs enhance scalability \cite{Asif_lcm}, enabling the handling of more extensive datasets and more complex tasks while setting new standards for efficiency and interpretability \cite{annirudhha_lcm, Simplify_LCM}.

Recognizing the comparatively limited academic research on LCMs, this study offers a comprehensive assessment of LCMs by synthesizing insights from grey literature, such as technical reports, blog posts, conference presentations, and YouTube discussions, which often provide early, practical perspectives on emerging technologies before formal peer-reviewed studies are available. This approach allows us to capture the latest developments and real-world implications of LCMs. Our analysis identifies the distinctive features that set LCMs apart from traditional LLMs, particularly their ability to reason at an abstract, language- and modality-agnostic level. It further examines their practical applications across critical domains such as cybersecurity, healthcare, and education while outlining key research directions and strategies for fostering their development and adoption. By synthesizing current knowledge, this study bridges the existing research gap, offering actionable insights for researchers and practitioners and emphasizing the pivotal role LCMs can play in shaping the next generation of interpretable, scalable, and context-aware AI systems.\\

In summary, our contributions are as follows:

\begin{itemize}

    \item \textbf{\textit{Identifying Distinctive Features}}: We identify the unique aspects that set LCMs apart from conventional LLMs, specifically their capacity to process information at a conceptual, language and modality-agnostic level.

    \item \textbf{\textit{Exploring Real-World Applications}}: We investigate the potential applications of LCMs across domains such as cybersecurity, healthcare, education, and others, demonstrating their ability to enhance contextual reasoning and deliver improved outcomes.

    \item \textbf{\textit{Providing LCM Implications}}: We offer future research avenues and practical recommendations for researchers and practitioners aimed at advancing the development, optimization, and adoption of LCMs.

\end{itemize}

\vspace{0.1in}

The structure of the paper is as follows: Section \ref{Section 2} discusses the conceptual workflow and architecture of LCMs. Section \ref{Section 3} discusses the research methodology used for our grey literature review. Section \ref{Section 4} presents and discusses our findings, followed by Section \ref{Section 5}, which examines the limitations of LCMs. Finally, Section \ref{Section 6} concludes the paper.

\section{Workflow and Architecture of LCMs} \label{Section 2}

This section presents the core design of LCMs, emphasizing how they process semantic units rather than individual tokens to improve long-context understanding and cross-modal reasoning.

\begin{figure}[t!]
    \centering
    \includegraphics[width=0.43\textwidth]{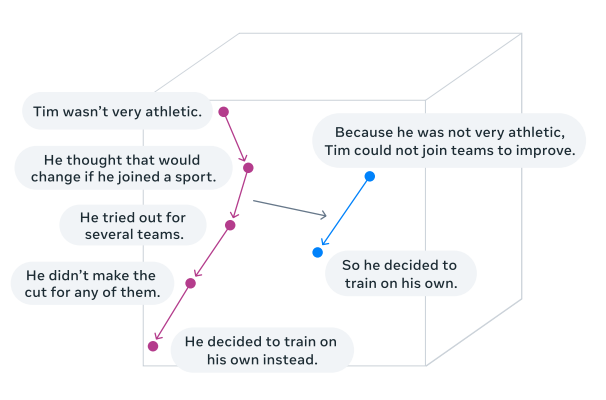}
    \caption{Visualization of LCMs' reasoning in an embedding space of concepts for summarization tasks \cite{the2024large}.}
    \label{workflow}
\end{figure}

\subsection{Conceptual Workflow of LCM}

Figure~\ref{workflow} depicts how LCMs handle input at a higher semantic level by reasoning in terms of \emph{concepts} rather than individual tokens \cite{Jiaxin_lcm}. Unlike LLMs, which predict the next word or token in a sequence, LCMs predict the next \emph{concept}, a complete thought, sentence, or idea \cite{Richard_lcm}. This conceptual shift enables the model to maintain both local context and global coherence, producing more meaningful and organized outputs \cite{Aniket_LCM}. In the example shown in Figure~\ref{workflow}, the LCM processes a story about a person’s sports journey. Each concept in the story represents a distinct yet interconnected idea within the narrative \cite{Siddhant_lcm}. For instance, the statements “Tim wasn’t very athletic” and “He tried out for several teams” share a close semantic relationship, reflected in their proximity in the embedding space. These concepts are encoded as vectors in a high-dimensional space, where semantically related ideas are positioned near each other \cite{Ganesh_LCM}. Drawing on these spatial relationships, the LCM predicts the next logical concept, such as “So he decided to train on his own,” demonstrating its ability to reason about sequences of ideas rather than merely guessing the next word \cite{Fahd_lcm, aii_LCM}.

By operating at the concept level, the LCM focuses on the “big picture” of the narrative rather than getting caught up in individual word predictions \cite{Wes_lcm, Leadership_LCM}. This holistic approach is especially beneficial for generating long-form content \cite{Anaam_lcm, BazAI_LCM}, where maintaining overall coherence and thematic continuity is essential. Concept-level reasoning allows the LCM to capture both short-term dependencies, such as the immediate context of a sentence, and long-term dependencies, such as the overarching structure and purpose of the text \cite{TheAIGRID_LCM}. This ensures the story retains a consistent narrative flow \cite{TheAILabsCanada_LCM}, which is particularly advantageous in applications where relationships between different sections of text are critical to grasping the intended meaning \cite{Discover_LCM, Next_LCM}.

\begin{figure}[t!]
    \centering
    \includegraphics[width=0.435\textwidth]{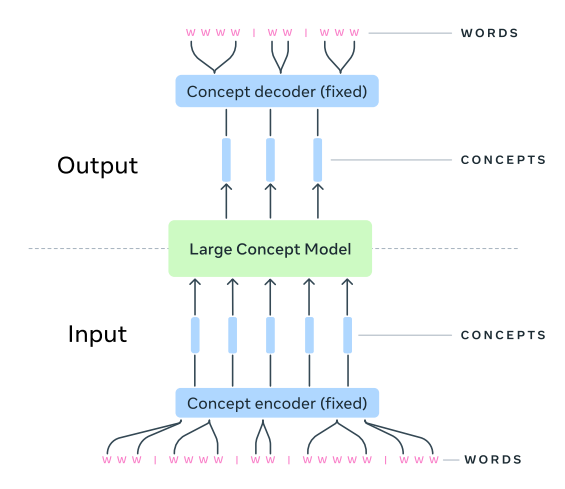}
    \caption{Fundamental architecture of Large Concept Model \cite{the2024large}.}
    \label{Architecture}
\end{figure}

\subsection{Architecture of the Large Concept Model}
Figure~\ref{Architecture} illustrates the architecture of LCM, composed of three primary components: the \emph{Concept Encoder}, the \emph{LCM Core}, and the \emph{Concept Decoder} \cite{Quantum_LCM, Teendifferent_LCM}. Working together, these components transform input into semantic embeddings, carry out high-level reasoning, and convert embeddings back into text or speech \cite{DataScience_lcm}. This cohesive architecture enables LCMs to generate contextually rich, coherent outputs across multiple languages and modalities \cite{Swaroop_lcm}.

\subsubsection{Concept Encoder}
The Concept Encoder translates sentences or phrases into fixed-size vector embeddings that capture their semantic meaning \cite{Albert_LCM, Anaam_lcm}. Unlike conventional encoders, it is \emph{modality-agnostic}, supporting text, speech, and potentially other input types such as images \cite{Wes_lcm}. Its key features include:

\begin{itemize}[leftmargin=*]
    \item \textit{Multilingual and Multimodal Capabilities}: The encoder is powered by SONAR embeddings and supports over 200 languages for text and 76 languages for speech, which can seamlessly process text and speech inputs by mapping them into the same embedding space \cite{Bhavik_lcm}.
    \item \textit{Unified Embedding Space}:  Diverse input formats (e.g., a written sentence versus its audio clip) are encoded into the same conceptual space \cite{Ganesh_LCM}. For instance, “The cat is hungry” in text and speech form map to the same concept vector.
\end{itemize}

By embedding multiple input types into a shared semantic space, the Concept Encoder allows the LCM to perform cross-modal reasoning without requiring format-specific retraining \cite{Ajith_lcm}. This design ensures accurate and consistent processing of varied inputs \cite{Swaroop_lcm}.\\

\begin{table*}[t!]
\centering
\caption{Data Sources for this Study}
\label{tab:greylit_sources}
\renewcommand{\arraystretch}{1.3}
\begin{tabular}{p{3.7cm}p{12.5cm}}
\hline
\textbf{Source} & \textbf{Description} \\ \hline
\textbf{Blog Posts} & Providing detailed insights and expert opinions on LCM design, performance, and applications. \\ \hline
\textbf{YouTube Videos} & Offering presentations, technical walkthroughs, and discussions from webinars and independent creators. \\ \hline
\textbf{Technical Reports} & Providing formal documentation detailing the architecture, performance, and use cases of LCMs. \\ \hline
\textbf{Informal communications} & Social media platforms such as Reddit, Twitter, and LinkedIn served as valuable sources for capturing informal yet informative discussions, user experiences, and early feedback from the broader AI community. \\ \hline
\end{tabular}
\end{table*}

\begin{table}[t!]
    \centering
    \caption{Inclusion and Exclusion Criteria.}
    \label{tab:criteria}
    \renewcommand{\arraystretch}{1.3}
    \begin{tabular}{p{8.5cm}}
        \hline
        \textbf{Inclusion Criteria} \\
        \hline
        \textit{I1}: Sources that directly address LCMs, their features, applications, or implications. \\
        \hline
        \textit{I2}: Sources selected irrespective of their publication date \\
        \hline
        \textbf{Exclusion Criteria} \\
        \hline
        \textit{E1}: Sources that are written in a language other than English \\
        \hline
        \textit{E2}: Sources that focus solely on traditional LLMs without any mention or comparison of LCMs \\
        \hline
        \textit{E3}: Literature that cannot be accessed through digital libraries, open repositories, or standard online searches via platforms like Google. \\
        \hline
    \end{tabular}
\end{table}

\begin{table*}[t!]
    \caption{Data extraction form of this study.} 
    \label{tab:extraction} 
    \centering
    \renewcommand{\arraystretch}{1.1}
    {\begin{tabularx}{\textwidth}{C{2em}C{8em}XC{6em}}
        \hline
        \textbf{ID} &  \textbf{Data Item} &  \textbf{Description} &  \textbf{Research Questions} \\ \hline
        D1 & LCM Distinctive Features & Key characteristics that differentiate LCMs from traditional LLMs & RQ1 \\ \hline
        D2 & Use Cases and Applications & Specific examples of domains and scenarios where LCMs are applied or have the potential to be used & RQ2 \\ \hline 
        D3 & Implications for Researchers & Insights on how LCMs can impact research advancements & RQ3 \\ \hline 
        D4 & Implications for Practitioners & Practical considerations for industry professionals & RQ3 \\ \hline 
        D5 & Limitations & Identified challenges and limitations of LCMs & - \\ \hline
    \end{tabularx}
    }
\end{table*}

\subsubsection{LCM Core}
The LCM Core serves as the model’s primary reasoning engine \cite{Siddhant_lcm}. It processes sequences of concept embeddings and predicts subsequent logical concepts in an autoregressive fashion \cite{Anaam_lcm}. Rather than guessing individual words, the LCM Core outputs embeddings that represent entire thoughts or ideas \cite{Keven_LCM}. Its core mechanisms include:

\begin{itemize}[leftmargin=*]
    \item \textit{Diffusion-Based Inference}: The LCM Core uses a denoising diffusion process to refine noisy intermediate embeddings \cite{AIPapers_lcm}. This iterative refinement step ensures that the predicted embeddings align closely with meaningful concepts by learning a conditional probability distribution over the embedding space \cite{Aniket_LCM}.
    \item \textit{Denoising Mechanism}: The diffusion process progressively removes noise from the predicted embeddings, making them more plausible and contextually relevant \cite{ AIPapers_lcm}.
    \item \textit{Hierarchical Reasoning}: The LCM Core models the progression of ideas across long contexts, maintaining narrative coherence and logical flow \cite{annirudhha_lcm}. 
\end{itemize}

This hierarchical design enables the LCM to anticipate upcoming concepts, resulting in outputs that are not just grammatically accurate but also contextually meaningful \cite{Zenspire7_lcm}.\\

\subsubsection{Concept Decoder}
The Concept Decoder transforms the refined embeddings generated by the LCM Core back into user-readable outputs, which can be text or speech \cite{ashish_lcm}. Its main features include:

\begin{itemize}[leftmargin=*]
    \item \textit{Reconstruction of Concepts}: The decoder converts abstract semantic embeddings into grammatically correct, semantically robust sentences, preserving the original intent \cite{Albert_LCM}.
    \item \textit{Cross-Modal Consistency}: Since the Concept Encoder and Decoder operate within the same embedding space, the LCM can seamlessly convert a single concept embedding into multiple formats \cite{Ganesh_LCM}. For example, the same concept vector can be decoded into different languages or spoken outputs.
\end{itemize}

The Concept Decoder ensures the final output remains faithful to the initial input while maintaining semantic clarity and fluency across languages and modalities \cite{Ajith_lcm}.

\section{Research Methodology}  \label{Section 3}

This section describes the methodology used to conduct a grey literature review for this study. Given the novelty of LCMs and the limited availability of peer-reviewed literature, a systematic approach was adopted to capture insights from grey literature sources. We followed a five-step process designed to ensure a comprehensive and rigorous review. These steps included defining research questions, identifying relevant sources, applying inclusion and exclusion criteria, data extraction, and synthesis of findings.

\subsection {Research Questions}

Defining clear research questions (RQs) is a critical step in guiding the direction of the study. The RQs formulated for this study are as follows:

\begin{itemize} [leftmargin=*]
    \item \textit{\textbf{RQ1}}. What are the key characteristics that distinguish LCMs from LLMs?
    \item \textbf{\textit{RQ2}}. What are the potential fields of application for LCMs?
    \item \textbf{\textit{RQ3}}. What are the implications of LCMs for researchers and practitioners in advancing innovation and practical adoption?
\end{itemize}

\vspace{0.1in}

These research questions address the primary objectives of the study: understanding the distinctive features of LCMs, identifying their practical use cases across various domains, and exploring their broader implications for research and real-world implementation. The research questions are designed to frame the study within a practical and theoretical context, ensuring that the investigation captures both the technical advancements introduced by LCMs and their potential impact across different sectors. By answering these questions, the study aims to provide valuable insights into the capabilities of LCMs, their cross-domain potential, and the challenges and opportunities associated with their adoption.

\subsection {Data Sources}

To answer the research questions comprehensively, we drew
from a wide range of grey literature sources. Table \ref{tab:greylit_sources} presents
the data sources used for the literature review for this study. By incorporating diverse sources of grey literature, we
aimed to capture a broad spectrum of perspectives, from formal
reports to community-driven insights. This approach ensures
that our findings reflect both technical advancements and
real-world implications of LCMs as discussed across various
platforms.
\subsection{Inclusion and Exclusion Criteria}

To ensure the reliability and relevance of the grey literature
reviewed, we applied a set of inclusion and exclusion criteria
during the selection process. Table \ref{tab:criteria} presents the inclusion
and exclusion criteria used for this study. These criteria helped
filter sources to focus on informative content directly related
to the RQs of the study. By applying these criteria, we ensured that the selected
literature provides a comprehensive, unbiased, and up-to-date
representation of LCM-related developments and their broader
implications

\begin{table*}[t!]
  \centering
  \caption{Comparison of LCMs and LLMs Across Various Characteristics}
  \label{tab:lcm_vs_llm}
  \renewcommand{\arraystretch}{1.3}
  \begin{tabular}{@{}p{3.5cm} p{7cm} p{6.5cm}@{}}
    \toprule
    \textbf{Characteristic} & \textbf{LCMs} & \textbf{LLMs} \\
    \midrule
    {Processing Unit} & 
      Sentences as concepts (semantic units). & 
      Individual tokens (words/subwords). \\ \hline
      
    {Reasoning Approach} & 
      Hierarchical, conceptual reasoning. & 
      Sequential, token-based reasoning. \\ \hline
      
    {Multilingual Support} & 
      Language-agnostic embeddings; supports 200+ languages. & 
      Requires fine-tuning for low-resource languages. \\ \hline

    {Long-Context Handling} & 
      Handles long documents efficiently. & 
      Computationally expensive for long texts. \\ \hline
      
    {Stability} & 
      Uses diffusion and quantization for robustness. & 
      Susceptible to inconsistencies in ambiguous inputs. \\ \hline
      
    {Generalization} & 
      Strong zero-shot generalization across tasks. & 
      Requires fine-tuning or large datasets for unseen tasks. \\ \hline
      
    {Architecture} & 
      Modular design (e.g., One-Tower, Two-Tower variants). & 
      Monolithic transformer-based architecture. \\ 
    \bottomrule
  \end{tabular}
\end{table*}

\subsection{Screening and Selection}

After gathering potential sources, each item underwent a careful evaluation based on the inclusion and exclusion criteria. This review involved manually verifying the alignment of source content with the RQs, noting aspects such as publication platform, author or organizational affiliation, and depth of discussion on LCM-related topics. Sources offering meaningful insights into LCM architecture, applications, and implications received priority. Materials deemed incomplete, promotional, or tangential to the RQs were excluded. When the relevance of a source was uncertain, a second review was conducted to ensure consistency and avoid bias. Ultimately, the final set of sources represented a diverse cross-section of the grey literature, including in-depth reports, community forums, and technical documentation.

\subsection{Data Extraction}

The data extraction process was designed to collect relevant information from the selected sources to address RQs. Key data points were identified and categorized based on their relevance to RQs. Table \ref{tab:extraction} presents the details of the data items (D1 to D5) included in the extraction process. By using this structured approach, relevant information from each source was categorized according to the data extraction form, ensuring that the findings addressed the research questions comprehensively. The extracted data was then synthesized to provide insights into the distinctive features, applications, and broader implications of LCMs, forming the foundation for the analysis and discussion of this study.

\section{Research Findings and Discussion} \label{Section 4}

This section presents the key findings of the study and provides an in-depth discussion aligned with the research questions. The analysis highlights the distinctive features, practical applications, and broader implications of LCMs, offering valuable insights for both researchers and practitioners.

\subsection{Distinctive Characteristics of LCMs}
\label{sec:distinctive_features}

This subsection addresses our first research question:

\begin{quote}
\textit{\textbf{RQ1}: What are the key characteristics that distinguish LCMs from LLMs?}
\end{quote}

The development of LCMs represents a substantial leap beyond traditional LLMs by moving from token-based to concept-level reasoning \cite{Siddhant_lcm}. This paradigm shift aims to alleviate known LLM limitations and enhance performance in areas such as coherence, multilingual adaptability, and structured text generation \cite{bjx_fun_lcm}. Table \ref{tab:lcm_vs_llm} provides a comparative overview of these two approaches, focusing on their differences across core characteristics. The following list details the unique features that set LCMs apart from LLMs.
\vspace{0.1in}

\begin{enumerate}[leftmargin=*]

\item \textbf{Processing Units - Concepts vs. Tokens:} LCMs operate at the sentence level, treating each sentence as a \emph{concept}, a self-contained semantic unit \cite{Income_lcm}. Instead of processing each word or token individually, LCMs encode entire sentences as conceptual vectors in a higher-dimensional semantic embedding space \cite{Ganesh_LCM}. For instance, when interpreting a historical event, LCMs focus on the broader meaning of a paragraph rather than getting bogged down by specific dates or details \cite{Anthony_LCM}. In contrast, LLMs handle input at the token level, processing one word or subword at a time \cite{Synced_lcm}. This fine-grained approach can lead to difficulties in maintaining coherence over long sequences, particularly for extended text generation.

\vspace{0.1in}
\begin{custombox}
{LCMs process fewer units (sentences  instead of  tokens), enabling them to handle large contexts more efficiently and produce more structured outputs, while LLMs focus on token-level precision.}
\end{custombox}

\vspace{0.1in}

\item \textbf{Reasoning and Abstraction Capabilities:} 
LCMs are intentionally designed for hierarchical reasoning and abstraction \cite{Swaroop_lcm}. By working at the sentence (concept) level, LCMs can form relationships among ideas and apply contextual reasoning \cite{Jiaxin_lcm}, much like humans linking concepts during a conversation. This structure enables LCMs to build a cohesive overview of the text rather than relying on the more granular, word-level relationships seen in LLMs \cite{Absaar_lcm}. While LLMs exhibit emergent abilities in summarization and translation, their reasoning is often implicit and acquired through large-scale token-based training.
\vspace{0.1in}

\begin{custombox}
{LCMs explicitly model relationships between semantic units, supporting more structured and human-like reasoning, whereas LLMs depend on token-based correlations and implicit pattern learning.}
\end{custombox}

\vspace{0.1in}

\item \textbf{Multilingual and Multimodal Support:} LCMs rely on the SONAR embedding space \cite{Synced_lcm}, a language-agnostic system that supports over 200 languages for text and 76 languages for speech, with experimental capabilities for sign language \cite{Bhavik_lcm}. This design allows LCMs to manage various languages seamlessly without the need for retraining \cite{Ganesh_LCM}. For example, an LCM can interpret an English document and generate a summary in Spanish using the same conceptual framework. In contrast, LLMs process text at the token level and often require language-specific vocabularies or tokenization strategies, limiting their generalizability. Consequently, fine-tuning or additional training is typically needed for LLMs to support low-resource languages or new modalities.

\vspace{0.1in}

\begin{custombox}
{LCMs inherently support multilingual and multimodal input/output, making them highly scalable across languages and formats. LLMs may require additional data or fine-tuning for cross-lingual or multimodal tasks.}
\end{custombox}
\vspace{0.1in}

\item \textbf{Long-Context Handling and Efficiency:} By processing inputs at the sentence level, LCMs maintain shorter overall sequence lengths compared to token-based models. This structure enables them to manage extensive contexts, such as lengthy reports, narratives, or documents, more efficiently \cite{szymonslowik_lcm}. Their architecture reduces computational overhead while preserving coherence across large spans of text \cite{viliotti_lcm_semantic_reasoning}. LLMs, however, process each token individually, leading to quadratic attention complexity and increased resource requirements for long-form content.

\vspace{0.1in}
\begin{custombox}
{LCMs can process long documents by encoding fewer conceptual units (sentences) rather than thousands of tokens, whereas LLMs require large memory and computation to handle long-form text due to their token-based processing.}
\end{custombox}
\vspace{0.1in}

\item \textbf{Stability and Robustness:} 
LCMs incorporate quantization and diffusion techniques to mitigate errors from minor input disturbances \cite{Albert_LCM, pruseth_lcm_agi}. Diffusion progressively refines noisy embeddings into coherent representations, while quantization converts continuous embeddings into discrete units, enhancing robustness against small deviations \cite{AIPapers_lcm}. In contrast, LLMs generally depend on optimized transformer architectures without explicit mechanisms for handling noisy or ambiguous inputs, leaving them more susceptible to inconsistencies or hallucinations.

\vspace{0.1in}

\begin{custombox}
{LCMs incorporate additional techniques like diffusion and quantization to stabilize outputs and improve robustness, whereas LLMs lack explicit mechanisms for handling noisy or ambiguous inputs.}
\end{custombox}
\vspace{0.1in}

\item \textbf{Zero-Shot Generalization:} LCMs demonstrate exceptional zero-shot generalization across various tasks, languages, and modalities by leveraging language-agnostic conceptual embeddings \cite{Asif_lcm}. This allows them to perform tasks such as summarization, translation, and content expansion in new languages or formats without requiring any additional training or fine-tuning \cite{AIbase_LCM}. For instance, an LCM can seamlessly read an English document and produce a detailed summary in Spanish by reasoning at the conceptual level, independent of language-specific features. This capability is largely attributed to the SONAR embedding space, which supports multilingual and multimodal input \cite{Ganesh_LCM}. In contrast, LLMs typically process input at the token level and rely on language-specific patterns learned during pretraining. As a result, they often require extensive fine-tuning or additional training data to generalize effectively to low-resource languages or unseen tasks, making them more dependent on the diversity of their training datasets and susceptible to performance limitations in novel contexts.

\vspace{0.1in}

\begin{custombox}
{LCMs can generalize across languages and tasks without retraining, while LLMs may require additional training or fine-tuning for similar performance.}
\end{custombox}
\vspace{0.1in}

\item \textbf{Architectural Modularity and Extensibility:} LCMs offer a highly modular design, supporting flexible architectures such as One-Tower and Two-Tower models \cite{DataScience_lcm}. The One-Tower model combines context processing and sentence generation in a single transformer, streamlining the workflow, while the Two-Tower model separates the context understanding phase from the generation phase, enhancing modularity and enabling more efficient specialization \cite{szymonslowik_lcm}. This design allows encoders and decoders to be developed or replaced independently, making it simpler to add support for new languages or modalities without significant architectural changes or retraining \cite{Ajith_lcm}.  In contrast, LLMs generally follow a monolithic transformer architecture where the entire model processes input from tokens to output, making modifications or extensions more complex and requiring extensive retraining to incorporate new capabilities. This limits their flexibility compared to the modular framework of LCMs, especially when adapting to new domains or tasks.

\vspace{0.1in}

\begin{custombox}
{LCMs’ modular architecture supports flexible extensions and independent updates to encoders and decoders, whereas LLMs are typically built as large, integrated models that require extensive retraining for updates. }

\end{custombox}

\end{enumerate}

Overall, these characteristics position LCMs as a powerful alternative to conventional LLMs. By addressing issues such as long-context coherence, cross-lingual support, and computational efficiency, LCMs pave the way for robust, human-like reasoning across an expansive range of use cases.

\begin{table*}[t!]
\centering
\caption{LCM Applications Across Key Domains}
\renewcommand{\arraystretch}{1.3}
\begin{tabularx}{\textwidth}{p{3.5cm}p{7.5cm}p{6cm}}
\hline
\textbf{Domain} & \textbf{Key Tasks} & \textbf{Potential Benefits} \\
\hline
{Multilingual NLP} & Cross-lingual Q\&A, multilingual content generation, translation/localization & Enhanced communication across languages, support for low-resource languages, real-time multilingual interactions \\
\hline
{Multimodal AI Systems} & Conversational AI, audio-visual processing, sign language interpretation & Unified multimodal integration, accessibility improvements, consistent user experience \\
\hline
{Healthcare and Medical} & Medical record insights, multilingual support for clinical tasks, research comparison & Faster diagnoses, improved patient communication, efficient research analysis \\
\hline
{Education and E-Learning} & Lesson extraction, feedback for language learners, essay evaluation & Personalized learning, accessible study materials, improved student performance \\
\hline
{Scientific Research and Collaboration} & Research synthesis, automated literature reviews, hypothesis generation & Faster knowledge aggregation, interdisciplinary insights, identification of research gaps \\
\hline
{Legal and Policy Analysis} & Policy comparison, legal content review, regulatory compliance checks & Reduced manual reviews, improved compliance, faster preparation for legal cases \\
\hline
{Human-AI Collaboration} & Writing assistance, collaborative authoring, advanced conversational agents & Increased productivity, refined content generation, improved thematic consistency \\
\hline
{Personalized Content Curation} & Streaming recommendations, tailored e-commerce suggestions, content customization & Improved user engagement, enhanced satisfaction, increased conversions \\
\hline

{Fraud Detection and Financial Analysis} & Fraud detection, financial report analysis, trend identification & Improved risk management, anomaly detection, faster insights from reports \\
\hline
{Cybersecurity and Threat Intelligence} & Threat pattern detection, automated incident responses & Faster mitigation of threats, enhanced detection accuracy, reduced false positives \\
\hline
{Financial Services and Risk Management} & Risk evaluations, portfolio optimization, market trend analysis & Proactive fraud prevention, diversified investments, accurate credit scoring \\
\hline
{Manufacturing and Supply Chain} & Workflow optimization, demand forecasting, disruption prediction & Lower operational costs, proactive issue resolution, resource optimization \\
\hline
{Retail and E-Commerce} & Personalized product recommendations, dynamic pricing adjustments & Higher sales, tailored user experience, increased conversion rates \\
\hline
{Transportation and Smart Cities} & Traffic flow management, route adjustments, public transit coordination & Minimized congestion, timely updates, efficient public services \\
\hline
{Public Safety and Emergency Response} & Crisis coordination, risk prediction, incident report analysis & Faster disaster response, effective resource allocation, enhanced preparedness \\
\hline
{Software Development} & Bug detection, requirement tracking, automatic documentation & Quicker debugging, better traceability, consistent documentation updates \\
\hline

\end{tabularx}
\label{tab:lcm_applications}
\end{table*}

\subsection{Applications of LCMs}

This subsection addresses our second research question:

\begin{quote}
\textit{\textbf{RQ2}: What are the potential fields of application for LCMs?}
\end{quote}
\vspace{0.05in}

The development of LCMs marks a significant advancement in natural language understanding and generation by moving from token-level processing to concept-level reasoning. This paradigm shift enables LCMs to excel in tasks that require coherence, multilingual adaptability, and structured generation \cite{Py_LCM}. Table \ref{tab:lcm_applications} presents an overview of the key applications of LCMs across various domains, showcasing their primary tasks and corresponding benefits. Below is a detailed exploration of the potential fields of application where LCMs can make a meaningful impact:

\vspace{0.1in}

\begin{enumerate}[leftmargin=*]

    \item \textbf{Multilingual Natural Language Processing:}  
LCMs excel at multilingual tasks due to their conceptual reasoning approach, which operates on language-agnostic embeddings rather than language-specific tokens \cite{Daniele_lcm}. By representing sentences as high-level concepts, LCMs can understand and generate content across multiple languages without requiring additional retraining. Their ability to reason using shared semantic structures, even when languages have distinct vocabularies and syntactic patterns, positions them as highly effective tools for breaking language barriers and facilitating cross-lingual communication.

\vspace{0.05in}

    \textbf{Applications:}
    \begin{itemize}[leftmargin=*]
        \item \textit{Multilingual Summarization:}  
        LCMs can summarize content written in one language (e.g., English) and generate summaries in another language (e.g., French) using the same conceptual reasoning process \cite{Asif_lcm}.\\
\textit{Use Case:} Global news agencies can create multilingual news briefs without building separate models for each language, making real-time reporting more inclusive and accessible.
        
        \item \textit{Cross-Lingual Question Answering:}  
        LCMs can answer questions posed in one language by retrieving relevant information stored in another language due to their ability to reason over language-independent concepts.\\
\textit{Use Case:} In multilingual customer support systems, an LCM-based chatbot can understand customer queries in Spanish and retrieve responses from an English knowledge base.
        
        \item \textit{Translation and Localization:}  
LCMs can handle translation and localization tasks involving low-resource languages that lack large training datasets, bridging the digital divide in underrepresented linguistic communities.\\
\textit{Use Case:} NGOs working in remote areas can use LCM-powered systems to translate emergency guidelines or health advisories into indigenous languages.

    \end{itemize}
    
    \begin{custombox} 
{The language-agnostic design of LCMs empowers complex multilingual tasks with minimal training, significantly enhancing global communication and collaboration. By unifying tasks like summarization, translation, and question-answering under a conceptual reasoning framework, LCMs set new benchmarks for multilingual NLP systems, fostering accessibility and inclusivity across diverse linguistic contexts.
}    \end{custombox}
    \vspace{0.1in}

\item \textbf{Multimodal AI Systems:}  
LCMs can handle diverse data formats such as text, speech, and experimental modalities like sign language by working with conceptual embeddings rather than language-specific tokens \cite{Income_lcm}. This unified approach enables LCMs to seamlessly process and integrate information across multiple modalities, fostering the development of inclusive and versatile AI systems capable of enhancing communication and understanding across different input formats \cite{AIbase_LCM}.

\vspace{0.05in}

\textbf{Applications:}
\begin{itemize}[leftmargin=*]
    \item \textit{Conversational AI:}  
    LCMs can build virtual assistants capable of understanding and responding to queries in various formats, including text, audio, and speech.\\
    \textit{Use Case:} Customer support systems can use LCMs to offer consistent service across chat, voice calls, and written messages, providing a unified and responsive user experience.
    
    \item \textit{Audio-Visual Summarization:}  
    LCMs can summarize multimedia content by processing speech transcripts and visual annotations together \cite{AIbase_LCM}.\\
    \textit{Use Case:} Educational platforms can provide text-based summaries of video lectures or podcasts, making content more accessible for users who prefer text-based formats or need a summary due to time constraints.
    
    \item \textit{Sign Language Translation:}  
    LCMs can facilitate real-time translation between sign language and spoken or written languages.\\
    \textit{Use Case:} Healthcare providers can use LCMs to assist in consultations with deaf or hard-of-hearing patients, ensuring accurate and inclusive communication in medical settings.
\end{itemize}

\begin{custombox} 
{LCMs maintain efficient resource allocation by operating on unified conceptual embeddings, enabling seamless integration of diverse input types. This capability supports the creation of more accessible, interactive, and inclusive AI systems that foster communication and understanding across different modalities, enhancing user experience.}
\end{custombox}
\vspace{0.1in}

\item \textbf{Healthcare and Medical:}  
LCMs can significantly enhance medical information processing by summarizing and contextualizing complex medical documents \cite{Zenspire7_lcm, TOPAI_LCM}. Their ability to maintain coherence across long texts enables effective summarization and comparison of patient histories, clinical reports, and research findings, streamlining documentation and supporting informed decision-making.

\vspace{0.05in}

\textbf{Applications:}
\begin{itemize}[leftmargin=*]
    \item \textit{Medical Record Summarization:}  
    LCMs can create concise summaries of patient histories for quick reference by healthcare providers.\\
    \textit{Use Case:} Physicians can review LCM-generated summaries of patient records during consultations, leading to faster and more informed diagnoses.
    
    \item \textit{Cross-Lingual Support:}  
    LCMs can translate medical instructions, discharge summaries, and reports into different languages, ensuring clear communication with patients from diverse backgrounds.\\
    \textit{Use Case:} Hospitals can provide multilingual post-discharge care instructions, improving adherence to medical advice and patient outcomes.
    
    \item \textit{Clinical Research:}  
    LCMs can summarize findings from clinical trial reports and compare outcomes across multiple studies.\\
    \textit{Use Case:} Research institutions can use LCMs to quickly analyze clinical trial results and identify trends or anomalies for further investigation.
\end{itemize}

\begin{custombox} 
{Medical documents are often lengthy and dense, making it challenging for healthcare providers to extract relevant information quickly. LCMs' ability to process long-form documents with precision and coherence reduces documentation burdens and improves patient care by providing clear, accessible, and accurate medical information.}
\end{custombox}
\vspace{0.1in}

\item \textbf{Education and E-Learning:}  
LCMs can enhance e-learning platforms by supporting personalized content generation and interactive educational experiences. Their sentence-level reasoning allows for precise feedback, customized lessons, and efficient content generation, making learning more engaging and accessible.

\vspace{0.05in}

\textbf{Applications:}
\begin{itemize}[leftmargin=*]
    \item \textit{Lesson Summarization:}  
    LCMs can generate study notes from textbooks, lecture transcripts, or online course materials.\\
    \textit{Use Case:} Students and educators can use LCM-generated summaries to quickly review key topics before exams or classroom discussions.
    
    \item \textit{Language Learning:}  
    LCMs can provide sentence-level corrections, explanations, and personalized feedback for language learners.\\
    \textit{Use Case:} Language learning apps can integrate LCMs to deliver grammar corrections, sentence-level translations, and contextual language practice.
    
    \item \textit{Essay Feedback:}  
    LCMs can offer detailed, sentence-level feedback on student essays by identifying conceptual errors and suggesting improvements.\\
    \textit{Use Case:} Educational platforms can use LCMs to provide constructive feedback on written assignments, helping students improve their writing and analytical skills.
\end{itemize}

\begin{custombox} 
{LCMs' multilingual support makes educational content more accessible to non-native speakers, while their conceptual reasoning capabilities provide precise feedback and personalized learning experiences. This enhances learning outcomes, student engagement, and overall academic performance.}
\end{custombox}
\vspace{0.1in}

\item \textbf{Cross-Domain Scientific Research and Collaboration:}  
LCMs can support interdisciplinary research by generalizing concepts across fields and summarizing technical literature \cite{Arman_LCM, Ajith_lcm}. By identifying connections between concepts, LCMs facilitate collaborative research and knowledge transfer across disciplines.

\vspace{0.05in}

\textbf{Applications:}
\begin{itemize}[leftmargin=*]
    \item \textit{Scientific Paper Summarization:}  
    LCMs can summarize complex research papers from multiple disciplines into concise, accessible summaries.\\
    \textit{Use Case:} Research institutions can streamline the review process by using LCMs to produce summaries of recent publications across various fields.
    
    \item \textit{Literature Review Automation:}  
    LCMs can generate reviews of related research articles, identifying key trends and gaps in the literature.\\
    \textit{Use Case:} Researchers can use LCMs to automate literature reviews and focus on analyzing results rather than compiling references.
    
    \item \textit{Hypothesis Generation:}  
    LCMs can assist researchers in formulating hypotheses by identifying patterns and connections between concepts across different studies.\\
    \textit{Use Case:} Cross-disciplinary research teams can leverage LCMs to identify new research directions by integrating insights from multiple domains.
\end{itemize}

\begin{custombox}
{Scientific research often requires synthesizing information from various fields and languages. LCMs' ability to break down language and domain barriers accelerates scientific discoveries and fosters collaboration, driving innovation and progress.}
\end{custombox}
\vspace{0.1in}

\item \textbf{Legal and Policy Analysis:}  
LCMs' ability to process sentence-level embeddings and maintain coherence over long contexts makes them highly effective for analyzing legal and policy documents \cite{Ryan_lcm}. By reasoning over conceptual representations, LCMs can quickly identify key points, compare policies, and detect relevant regulatory clauses, improving efficiency in legal and policy workflows.

\vspace{0.05in}

\textbf{Applications:}
\begin{itemize}[leftmargin=*]
    \item \textit{Policy Comparison:}  
    Compare policies from different jurisdictions or languages to identify differences and similarities.\\
    \textit{Use Case:} Government agencies can use LCMs to compare environmental regulations across multiple regions and draft uniform policy proposals based on commonalities.
    
    \item \textit{Legal Summarization:}  
    Summarize lengthy court rulings, contracts, or legal precedents into concise, structured briefs.\\
    \textit{Use Case:} Legal professionals can save time during case preparations by accessing LCM-generated summaries of precedents, reducing the need for exhaustive manual reviews.
    
    \item \textit{Regulatory Compliance:}  
    Automatically identify and extract regulatory clauses from lengthy legal documents to ensure compliance with laws and policies \cite{Ryan_lcm}.\\
    \textit{Use Case:} Corporations can use LCM-powered tools to monitor changes in regulations and update their compliance frameworks efficiently.
\end{itemize}

\begin{custombox}
{Legal and policy documents often span hundreds of pages and contain complex language.  LCMs excel at long-context processing, enabling legal professionals to quickly extract relevant insights and focus on higher-value analysis, such as legal strategy and case development.}
\end{custombox}
\vspace{0.1in}

\item \textbf{Human-AI Collaboration and Interactive Systems:}  
LCMs' modular architecture and transparent decision-making capabilities make them ideal for interactive AI systems that require high interpretability and collaboration \cite{Daniele_lcm}. By generating coherent and explainable outputs, LCMs can enhance user trust and facilitate more effective human-AI partnerships in content creation and decision-making processes.

\vspace{0.05in}

\textbf{Applications:}
\begin{itemize}[leftmargin=*]
    \item \textit{Interactive Writing Tools:}  
    Provide high-level suggestions and corrections to improve the overall coherence, style, and structure of written content.\\
    \textit{Use Case:} Academics and professionals can use LCM-based writing assistants to enhance their research papers, proposals, or presentations, ensuring logical flow and thematic consistency.
    
    \item \textit{Conversational Agents:}  
    Build chatbots capable of generating detailed, context-aware answers based on conceptual understanding.\\
    \textit{Use Case:} Customer service platforms can implement LCM-powered virtual assistants that provide in-depth answers to user queries across multiple languages and formats.
    
    \item \textit{Collaborative Content Creation:}  
    Assist users in co-writing documents by maintaining thematic consistency, suggesting improvements, and tracking dependencies between sections \cite{Daniele_lcm}.\\
    \textit{Use Case:} News agencies can use LCMs to co-author breaking news stories in real time by summarizing events and refining details collaboratively.
\end{itemize}

\begin{custombox}
{Effective human-AI collaboration requires interpretability, adaptability, and trust. LCMs' conceptual reasoning provides transparency and consistency, enabling users to refine outputs iteratively. This fosters a more interactive and user-centric approach to AI-assisted workflows, making LCMs valuable for content creators, researchers, and decision-makers across various domains.}
\end{custombox}
\vspace{0.1in}

\item \textbf{Personalized Recommendations and Content Curation:}  
LCMs can enhance recommendation systems by reasoning over user preferences and conceptual relationships between content items. Unlike traditional systems that match keywords or numerical scores, LCMs can understand the thematic connections between content, resulting in more personalized and context-aware recommendations.

\vspace{0.05in}

\textbf{Applications:}
\begin{itemize}[leftmargin=*]
    \item \textit{Media Streaming:}  
    Suggest movies, TV shows, or songs by identifying thematic similarities (e.g., recommending shows with similar narrative arcs).\\
    \textit{Use Case:} Streaming platforms like Netflix can use LCMs to recommend content based on user watch history and preferences, even across different languages or genres.
    
    \item \textit{E-Commerce Recommendations:}  
    Curate personalized product suggestions by reasoning over user reviews and product features.\\
    \textit{Use Case:} Online retailers can enhance cross-category recommendations by matching conceptually related products (e.g., ``outdoor camping gear" rather than just ``tent accessories").
\end{itemize}

\begin{custombox} 
{LCMs’ ability to understand thematic relationships and user preferences enables more accurate and context-aware recommendations. This improves user satisfaction and engagement across platforms, such as media streaming services and e-commerce websites.}
\end{custombox}
\vspace{0.1in}

\item \textbf{Fraud Detection and Financial Analysis:}  
LCMs can enhance the detection of fraudulent activities and improve financial document summarization by identifying semantic anomalies and patterns in transactions.

\vspace{0.05in}

\textbf{Applications:}
\begin{itemize}[leftmargin=*]
    \item \textit{Fraudulent Activity Detection:}  
    Identify discrepancies in transaction narratives or insurance claims.\\
    \textit{Use Case:} Banks and insurance companies can use LCMs to detect unusual patterns in customer claims or financial reports that traditional methods may miss.
    
    \item \textit{Financial Forecast Summarization:}  
    Summarize long financial reports and earnings statements into actionable insights.\\
    \textit{Use Case:} Investors and analysts can use LCMs to condense quarterly earnings reports into summaries for faster decision-making.
\end{itemize}

\begin{custombox} 
{LCMs' ability to detect semantic anomalies and summarize complex financial documents enables more efficient fraud detection and financial analysis, improving risk management and decision-making for financial institutions.}
\end{custombox}
\vspace{0.1in}

\item \textbf{Cybersecurity and Threat Intelligence:} 

LCMs have the potential to support a wide range of cybersecurity tasks, including risk assessment \cite{abdulsatar2024towards}, cyber situational awareness \cite{ahmad2024survey}, and vulnerability analysis \cite{jayalath2024microservice}. They can further enhance threat detection \cite{goel2024machine}, access control \cite{ahmad2023review}, and vulnerability management \cite{syed2020cybersecurity} by reasoning over security logs, attack patterns, and system configurations at a conceptual level, enabling more accurate and context-aware decision-making. By identifying semantic patterns and correlating diverse data sources, LCMs can provide real-time insights into potential security threats and recommend mitigation strategies.

\vspace{0.05in}

\textbf{Applications:}  
\begin{itemize}[leftmargin=*]  

    \item \textit{Threat Detection and Correlation:}  
    LCMs can detect and correlate sophisticated attack patterns by analyzing relationships between security events, even when spread across different systems and logs.\\  
    \textit{Use Case:} Security Operation Centers (SOCs) can leverage LCMs to detect advanced persistent threats (APTs) by correlating information from system logs, user behaviors, and network anomalies to identify complex attack chains that traditional tools may miss.  

    \item \textit{Incident Response Automation:}  
    LCMs can generate real-time incident summaries and suggest mitigation actions tailored to the context of the attack.\\  
    \textit{Use Case:} Cybersecurity teams can use LCMs to automatically create incident reports summarizing key attack details and recommend customized mitigation actions based on the specific vulnerabilities and system configurations affected.  

\end{itemize}  

\begin{custombox}  
{LCMs improve cybersecurity operations by correlating complex threat narratives, enabling faster threat detection, streamlined incident response, and more effective vulnerability management.}  
\end{custombox}  
\vspace{0.1in}

\item \textbf{Financial Services and Risk Management:}  
LCMs improve decision-making in financial services by identifying patterns and trends in complex datasets, enabling more accurate risk assessments, fraud detection, and portfolio optimization. By reasoning over interconnected financial data, LCMs can support institutions in making informed, data-driven decisions.

\vspace{0.05in}

\textbf{Applications:}  
\begin{itemize}[leftmargin=*]  

    \item \textit{Risk Assessment and Credit Analysis:}  
    LCMs can analyze market reports, customer profiles, and historical transaction data to assess financial risks more comprehensively.\\  
    \textit{Use Case:} Banks can use LCMs to evaluate loan applications by identifying hidden trends and relationships within financial histories and credit reports, improving the accuracy of credit scoring and risk prediction.  

    \item \textit{Portfolio Optimization and Investment Strategies:}  
    LCMs can identify thematic investment opportunities by linking economic reports, news sentiment, and market trends across multiple domains.\\  
    \textit{Use Case:} Investment firms can use LCMs to construct optimized portfolios by reasoning over financial news, company performance reports, and historical price data, enabling more effective diversification and returns optimization.  

\end{itemize}  

\begin{custombox}  
{LCMs enhance financial services by synthesizing data from multiple sources, enabling more accurate risk assessments, improved portfolio management, and better fraud detection, leading to more robust financial decision-making.}  
\end{custombox}  
\vspace{0.1in}

\item \textbf{Manufacturing and Supply Chain Optimization:}  
LCMs can optimize production workflows and supply chain logistics by reasoning over data from different stages of the manufacturing and distribution process. By correlating operational data, LCMs can enhance visibility across the supply chain, helping organizations identify inefficiencies and respond to disruptions proactively.

\vspace{0.05in}

\textbf{Applications:}  
\begin{itemize}[leftmargin=*]  

    \item \textit{Production Optimization:}  
    LCMs can identify inefficiencies in production lines and recommend improvements by reasoning over data from various production stages.\\  
    \textit{Use Case:} Manufacturing units can use LCMs to detect real-time bottlenecks, such as delays in assembly lines, and suggest optimal resource allocation strategies to improve throughput and minimize downtime.  

    \item \textit{Supply Chain Forecasting and Resilience:}  
    LCMs can improve logistics by correlating data from suppliers, demand forecasts, and transportation schedules to predict potential disruptions and recommend contingency plans.\\  
    \textit{Use Case:} Supply chain managers can use LCMs to identify risks in the distribution network—such as late shipments due to port closures—and adjust delivery routes or inventory levels accordingly.  

\end{itemize}  

\begin{custombox}  
{LCMs enhance supply chain resilience by synthesizing data from multiple stages to optimize resource utilization, improve forecasting accuracy, and minimize delays, contributing to smoother and more efficient operations.}  
\end{custombox}  
\vspace{0.1in}

\item \textbf{Personalized Retail and E-Commerce Experiences:}  
LCMs elevate customer engagement and optimize sales strategies by reasoning over consumer behavior, purchase history, and market trends. By understanding conceptual relationships between products and user preferences, LCMs can deliver more personalized and context-aware recommendations.

\vspace{0.05in}

\textbf{Applications:}  
\begin{itemize}[leftmargin=*]  

    \item \textit{Personalized Product Recommendations:}  
    Suggest products based on conceptual connections between user preferences and product categories.\\  
    \textit{Use Case:} E-commerce platforms can use LCMs to recommend curated product bundles (e.g., "home office essentials" rather than individual desks and chairs) to increase cart size and improve customer experience.  

    \item \textit{Dynamic Pricing Strategies:}  
    Optimize prices by reasoning over market demand, competitor pricing, and seasonal trends.\\  
    \textit{Use Case:} Retailers can use LCMs to implement real-time dynamic pricing strategies, adjusting prices during flash sales or holiday events to maximize revenue and customer engagement.  

\end{itemize}  

\begin{custombox}  
LCMs elevate both the customer experience and sales outcomes by offering tailored recommendations based on context and dynamic pricing models. These strategies foster higher levels of customer satisfaction and boost conversion rates, ultimately leading to increased sales performance.

\end{custombox}  
\vspace{0.1in}

\item \textbf{Smart Transportation and Urban Planning:}  
LCMs optimize traffic management and public transportation systems by reasoning over data from sensors, vehicles, and urban infrastructure \cite{Arman_LCM}. By identifying patterns in traffic flow and public transit usage, LCMs can enable smarter urban planning and real-time adjustments to improve mobility.

\vspace{0.05in}

\textbf{Applications:}  
\begin{itemize}[leftmargin=*]  

    \item \textit{Traffic Management and Congestion Control:}  
    Identify congestion points and suggest alternate routes by reasoning over real-time traffic data.\\  
    \textit{Use Case:} Smart city platforms can use LCMs to adjust traffic signal timings dynamically and reroute vehicles to ease congestion during peak hours or in response to road closures.  

    \item \textit{Public Transit Scheduling and Optimization:}  
    Adjust transit schedules by reasoning over passenger demand, operational constraints, and special events.\\  
    \textit{Use Case:} Transportation authorities can use LCMs to reschedule buses and trains during major events or emergencies to ensure efficient crowd management and minimize waiting times for passengers.  

\end{itemize}  

\begin{custombox}  
{LCMs improve transportation efficiency by processing and correlating real-time data, enabling better traffic flow, more reliable public transit schedules, and smarter resource allocation for urban mobility.}  
\end{custombox}  
\vspace{0.1in}

\item \textbf{Public Safety and Emergency Response:}  
LCMs enhance public safety and emergency response by synthesizing diverse data sources, such as emergency reports, environmental data, and public communications, to provide real-time insights and support informed decision-making. By identifying conceptual patterns and relationships across reports and contextual data, LCMs can improve crisis management, enabling faster and more effective resource allocation during emergencies.

\vspace{0.05in}

\textbf{Applications:}  
\begin{itemize}[leftmargin=*]  

    \item \textit{Disaster Response Coordination:}  
    LCMs can consolidate reports from various emergency services and environmental monitoring systems into concise, actionable summaries, prioritizing critical areas that need immediate attention.\\  
    \textit{Use Case:} Emergency response units can use LCMs to integrate data from first responders, meteorological services, and social media posts to create a holistic overview during disasters, such as hurricanes or earthquakes \cite{Arman_LCM}. This helps responders identify the most affected areas, streamline rescue operations, and allocate medical and logistical resources efficiently.  

    \item \textit{Predictive Risk Identification:}  
    LCMs can analyze historical disaster data, live environmental readings, and social media signals to predict potential threats and issue early warnings.\\  
    \textit{Use Case:} Public safety departments can use LCMs to anticipate incidents such as wildfires, floods, or chemical spills by reasoning over factors like weather forecasts, vegetation conditions, and industrial activity. By identifying areas of concern early, LCMs enable authorities to issue evacuation alerts or deploy preventive measures, such as firebreaks and sandbag reinforcements, before a crisis unfolds.  

    \item \textit{Incident Communication Management:}  
    LCMs can process and summarize incoming public communications, such as emergency calls, citizen reports, and social media updates, to detect patterns of distress and improve response coordination.\\  
    \textit{Use Case:} Emergency hotlines can leverage LCMs to categorize incoming calls, highlight recurring issues, and escalate high-priority incidents based on urgency, ensuring critical reports are addressed promptly and systematically.  

\end{itemize}  

\begin{custombox}  
{LCMs’ ability to process and correlate information from multiple sources improves public safety operations by enabling faster, data-driven decision-making and more efficient resource coordination during emergencies. Their predictive capabilities also enhance preparedness, mitigating the impact of potential disasters.}  
\end{custombox}  
\vspace{0.1in}

\item \textbf{Enhancing Software Development and Engineering Processes:}  

With the rise of modern and complex software architectures, such as microservices \cite{ahmad2024towards, Ahmad_2024}, advancing practices to improve the software development lifecycle has become crucial. LCMs can enhance software engineering processes by reasoning over source code, technical documentation, and system dependencies. This capability enables more effective bug detection, code optimization, and traceability between requirements and implementation, resulting in streamlined development workflows and higher-quality software.

\vspace{0.05in}

\textbf{Applications:}  
\begin{itemize}[leftmargin=*]  

    \item \textit{Semantic Code Reviews and Bug Detection:}  
    LCMs can detect inconsistencies, errors, and potential security vulnerabilities by reasoning over code semantics and identifying conceptual mismatches in function usage and dependencies.\\  
    \textit{Use Case:} Development teams can use LCMs during code reviews to detect logic flaws, such as null reference errors or race conditions, that conventional tools may miss. LCMs can also suggest optimized refactoring strategies to enhance code readability and maintainability.  

    \item \textit{Requirement Traceability and Impact Analysis:}  
    LCMs can link software requirements to corresponding code components and documentation, enabling better traceability and impact analysis during updates and feature expansions.\\  
    \textit{Use Case:} Software architects can use LCMs to track dependencies between user stories and code implementations, ensuring that changes to core modules do not inadvertently break related features. This is especially valuable in large-scale projects where multiple teams contribute to interconnected components.  

    \item \textit{Automated Documentation Generation:}  
    LCMs can generate comprehensive and context-aware documentation by reasoning over the purpose and interactions of different code segments.\\  
    \textit{Use Case:} Development teams can use LCMs to automatically create API documentation, design descriptions, and user guides that remain up-to-date as the code evolves, reducing the burden of manual documentation tasks.  

\end{itemize}  

\begin{custombox}  
{By automating code analysis and improving traceability, LCMs enhance software development processes, enabling faster bug detection, better maintenance practices, and more robust software solutions. Additionally, their ability to generate contextual documentation improves knowledge sharing across development teams.}  
\end{custombox}  
\vspace{0.1in}

   
\end{enumerate}

LCMs offer broad applicability across a wide range of industries, including healthcare, legal services, education, scientific research, and more. By leveraging sentence-level reasoning and conceptual embeddings, LCMs excel at handling long contexts, integrating multimodal inputs, and supporting multilingual communication \cite{viliotti_lcm_semantic_reasoning}. Their capacity to perform cross-lingual tasks and reason over complex information enables more accurate insights, informed decision-making, and streamlined workflows. These capabilities position LCMs as transformative tools for enhancing collaboration, innovation, and productivity across diverse domains, addressing challenges that traditional token-based models often struggle to overcome.\\

\begin{table*}[t!]
\centering
\caption{Implications for Researchers with LCMs}
\renewcommand{\arraystretch}{1.3}
\begin{tabular}{p{4cm}p{13cm}} 
\hline
\textbf{Key Dimension} & \textbf{What Researchers Can Do} \\
\hline
Redefining NLP Frameworks & Develop concept-level NLP models to improve coherence, create benchmarks for semantic reasoning, and explore long-context processing. \\
\hline
Interdisciplinary Research & Generate cross-domain hypotheses, build knowledge graphs linking different research fields, and foster interdisciplinary studies. \\
\hline
Semantic Representations & Design sentence-level embeddings and hierarchical models that link research findings to hypotheses and conclusions. \\
\hline
Explainability and Ethical AI & Create transparent models by using concept maps and summaries, improve bias detection, and enhance model interpretability. \\
\hline
New Research Frontiers & Investigate ambiguous language, idiomatic expressions, and develop models for low-resource languages and cultural nuances. \\
\hline
Multimodal Reasoning & Conduct studies on cross-modal learning using text, images, and audio to advance multimedia understanding and retrieval. \\
\hline
Collaborative Knowledge Bases & Build open-access repositories for sharing research insights and fostering collaboration across disciplines. \\
\hline
Real-Time Applications & Adapt LCMs for live transcription, multilingual speech-to-text systems, and real-time event summarization. \\
\hline
\end{tabular}
\label{tab:lcm_research}
\end{table*}

\subsection{Implications of LCMs}

This subsection addresses our third research question:

\begin{quote}
\textit{\textbf{RQ3}: What are the implications of LCMs for researchers and practitioners in advancing innovation and practical adoption?}
\end{quote}

\vspace{0.05in}

\noindent\textbf{Implications for Researchers.} LCMs provide researchers with substantial opportunities to revolutionize NLP frameworks and promote interdisciplinary advancements. By harnessing their conceptual-level reasoning, researchers can effectively support multilingual, multimodal, and long-context tasks, thereby expanding the horizons of NLP research. Table \ref{tab:lcm_research} highlights key areas where researchers can leverage LCMs across various domains. Below, we delve into an in-depth exploration of how LCMs can drive innovation and enable new research initiatives:

\vspace{0.1in}

    \begin{enumerate}[leftmargin=*]
    \item \textit{Redefining NLP Frameworks:} LCMs empower researchers to develop NLP systems that reason at a conceptual level, fostering more coherent and contextually aware language models.
    \begin{itemize}[leftmargin=*]
        \item {Conceptual Shift:} Researchers can transition from traditional token-based representations to capturing broader semantic connections in long-form texts by representing sentences as unified semantic units. This shift enhances the overall quality of language understanding and generation \cite{Ganesh_LCM, Income_lcm}.
        \item {New Benchmarks:} The adoption of concept-level reasoning encourages the creation of new evaluation benchmarks that focus on semantic reasoning, long-context generation, and cross-lingual capabilities, thereby advancing the standards of NLP research.
        \item {Advanced Architectures:} Researchers have the opportunity to experiment with hybrid NLP architectures that integrate sentence-level embeddings with traditional token-based models, bridging the gap between conceptual reasoning and token-level precision.
    \end{itemize}
    
    \vspace{0.05in}
    
    \item \textit{Interdisciplinary Research Advancements:} LCMs facilitate interdisciplinary collaborations by generalizing across languages and domains, fostering connections between fields such as NLP, cognitive science, education, and policy research \cite{Arman_LCM}.
    \begin{itemize}[leftmargin=*]
        \item {Cross-Domain Hypothesis Generation:} By identifying conceptual connections across various fields, researchers can use LCMs to generate hypotheses that span multiple disciplines, such as linking medical research findings to policy decisions \cite{sox2024medical} and data dimensionality reduction \cite{abbas2024robust}.
        \item {Knowledge Graph Construction:} LCMs enable the construction of knowledge graphs that map concepts across different domains, allowing for deeper exploration of how ideas are related across seemingly disparate fields.
        \item {Collaborative Insights:} Researchers can build models that support interdisciplinary collaboration by leveraging LCMs' semantic understanding of research articles, reports, and case studies across different languages and terminologies.
    \end{itemize}
    
    \vspace{0.05in}
    
\begin{table*}[t!]
\centering
\caption{Implications for Practitioners with LCMs}
\renewcommand{\arraystretch}{1.3}
\begin{tabular}{p{4cm}p{13cm}} 
\hline
\textbf{Key Dimension} & \textbf{What Practitioners Can Do} \\
\hline
Streamlined Workflows & Automate documentation and report generation to boost productivity in legal, financial, and healthcare operations. \\
\hline
Cross-Lingual and Multimodal Solutions & Implement virtual assistants that handle text, speech, and visual inputs to enhance multilingual and multimodal communication. \\
\hline
User Accessibility & Personalize e-learning platforms with tailored recommendations, localized content, and inclusive communication tools. \\
\hline
Regulatory Compliance & Automate regulatory tracking, compare policies, and generate legal document summaries to ensure compliance. \\
\hline
Medical Information Processing & Summarize medical records and translate key documents for improved cross-lingual care and faster clinical decisions. \\
\hline
Content Generation & Co-author reports, expand content outlines, and provide structured feedback to streamline collaborative writing. \\
\hline
Knowledge Management & Create annotated knowledge bases, improve semantic search, and link related reports for efficient information retrieval. \\
\hline
Customer Interaction & Develop multilingual, context-aware chatbots with real-time speech-to-text features for seamless customer engagement. \\
\hline
E-Learning and Training & Design adaptive learning modules, deliver real-time feedback, and identify learning gaps to enhance training outcomes. \\
\hline
Decision Support & Use LCM-generated insights for financial planning, and market research to make data-driven decisions. \\
\hline
\end{tabular}
\label{tab:lcm_practitioners}
\end{table*}

    \item \textit{Innovations in Semantic Representation:}  
    LCMs offer researchers the ability to develop richer semantic representations that capture complex relationships between ideas and concepts \cite{Jiaxin_lcm}.
    \begin{itemize}[leftmargin=*]
        \item {Rich Sentence-Level Embeddings:} Researchers can shift from focusing on individual words or phrases to representing entire sentences or paragraphs as unified semantic units, enabling deeper insights into meaning and structure.
        \item {Specialized Conceptual Models:} Researchers can create domain-specific conceptual embeddings tailored to legal, medical, or technical texts, thereby enhancing the performance of NLP systems in specialized fields.
        \item {Hierarchical Context Modeling:} Utilizing LCMs' sentence-level reasoning, researchers can model hierarchical dependencies between sections of a document, such as linking experimental results back to hypotheses or conclusions in research papers \cite{Zenspire7_lcm}.
    \end{itemize}
    
    \vspace{0.05in}
    
    \item \textit{Enhancing Explainability and Ethical AI:}  
    LCMs empower researchers to advance Explainable AI (XAI) initiatives by producing clearer, more interpretable outputs.
    \begin{itemize}[leftmargin=*]
        \item {Improved Transparency:} Researchers can design LCMs to present outputs as structured summaries or concept maps, allowing users to understand the decision-making process and enhancing model interpretability.
        \item {Alignment with Ethical Principles:} Researchers can refine LCMs to ensure that outputs adhere to ethical guidelines and regulatory requirements, such as fairness, bias detection, and privacy protection.
        \item {Error Diagnosis:} By leveraging LCMs' ability to highlight sentence-level conceptual flows, researchers can more effectively diagnose errors by pinpointing where a model's reasoning may have deviated from expected patterns.
    \end{itemize}
    
    \vspace{0.05in}
    
    \item \textit{New Research Frontiers:}  
    LCMs open up new avenues for research in understanding ambiguous language, idiomatic expressions, and contextual nuances across different languages and data formats.
    \begin{itemize}[leftmargin=*]
        \item {Handling Ambiguity:} Researchers can utilize LCMs to study how meaning changes based on context, helping to build models that understand and interpret ambiguous sentences or culturally specific phrases.
        \item {Cross-Lingual Contextualization:} By leveraging shared semantic structures, researchers can enhance language models' performance in low-resource languages and develop more effective translation tools.
        \item {Domain-Specific Optimizations:} Researchers can investigate how LCMs can be optimized for highly specialized tasks, such as legal or medical analysis, without sacrificing generalizability.
    \end{itemize}
    
    \vspace{0.05in}

    \item \textit{Improved Multimodal Reasoning:}  
    LCMs enable the processing of data from various modalities, such as text, speech, and sign language, making them ideal for exploring cross-modal reasoning.
    \begin{itemize}[leftmargin=*]
        \item {Multimodal Summarization:} By combining information from different data types (e.g., text and audio), researchers can use LCMs to generate summaries that capture a more comprehensive picture of an event or presentation.
        \item {Cross-Media Retrieval:} LCMs allow for the development of systems that retrieve related information across different media formats, such as finding relevant video segments based on textual queries.
        \item {Multimodal Representation Learning:} Researchers can leverage LCMs to learn joint representations of text, images, and audio, enabling deeper analysis of multimedia content.
    \end{itemize}
    
    \vspace{0.05in}
    
    \item \textit{Collaborative Knowledge Bases and Open Science:}  
    By establishing semantic links across various fields, LCMs contribute to the development of open, collaborative knowledge repositories.
    \begin{itemize}[leftmargin=*]
        \item {Semantic Linking Across Disciplines:} LCMs can map research findings from different domains, allowing researchers to identify thematic overlaps and collaborative opportunities.
        \item {Open Science Initiatives:} Researchers can utilize LCMs to organize and share research findings in structured formats, facilitating open-access research and cross-disciplinary studies.
        \item {Dynamic Research Insights:} By continuously updating conceptual links as new findings emerge, LCMs support the creation of living knowledge bases that evolve with scientific progress.
    \end{itemize}
    
    \vspace{0.05in}
    
    \item \textit{Adapting LCMs for Real-Time Use Cases:}  
    Researchers may explore ways to adapt LCMs for real-time applications, such as live transcription and multilingual speech-to-text.
    \begin{itemize}[leftmargin=*]
        \item {Live Transcription:} LCMs can be adapted for real-time transcription services that summarize ongoing events, such as conferences or court hearings, into concise and understandable notes.
        \item {Multilingual Speech-to-Text:} By converting speech in multiple languages into text, LCMs facilitate real-time communication and help break language barriers.
        \item {Event Summarization:} LCMs can be employed to summarize live news updates or sports events, enabling faster reporting and analysis for journalists and broadcasters.
    \end{itemize}
    
    \vspace{0.05in}
\end{enumerate}

\begin{custombox}
LCMs can drive innovation by enabling conceptual-level reasoning, fostering interdisciplinary collaboration, and enhancing multimodal analysis. They may support the development of new benchmarks, facilitate domain-specific optimizations, and promote ethical AI through improved transparency and interpretability. By integrating LCMs into real-time applications and knowledge repositories, researchers can advance open science initiatives and develop more impactful, context-aware AI systems.
\end{custombox}

\vspace{0.1in}
\noindent\textbf{Implications for Practitioners.} Practitioners can leverage LCMs to enhance interpretability, make informed decisions, and effectively tackle complex real-world challenges. By applying LCMs in their workflows, practitioners can improve efficiency, facilitate better communication, and develop robust solutions within their respective fields. Table \ref{tab:lcm_practitioners} outlines key areas where practitioners can utilize LCMs to achieve these objectives.

\begin{enumerate}[leftmargin=*]

    \item \textit{Streamlined Workflows and Automation:}  
    Practitioners can utilize LCMs to automate labor-intensive tasks, thereby improving productivity and freeing up time for higher-value activities.  
    \begin{itemize}[leftmargin=*]
        \item {Automated Documentation:} LCMs can summarize legal contracts, financial reports, and meeting transcripts, ensuring that key information is captured without manual intervention.
        \item {Enhanced Report Generation:} In industries such as healthcare and finance, practitioners can rely on LCMs to generate detailed yet concise reports from structured data inputs.
        \item {Task Prioritization:} By automating routine documentation, professionals can focus on strategy development, client engagement, and decision-making.
    \end{itemize}
    \vspace{0.05in}

    \item \textit{Cross-Lingual and Multimodal Solutions:}  
    LCMs support cross-lingual tasks and multimodal data integration, making applications more versatile and user-friendly.  
    \begin{itemize}[leftmargin=*]
        \item {Multilingual Content Creation:} Practitioners can generate content in multiple languages without additional retraining, enhancing global communication efforts.
        \item {Unified Virtual Assistants:} The multimodal capabilities of LCMs allow for the development of virtual assistants that support text, speech, and visual data, enabling seamless interactions across different input formats.
        \item {Consistent Customer Support:} Organizations can provide consistent support across languages and modalities using LCMs, thereby improving user experience in global customer engagement.
    \end{itemize}
    \vspace{0.05in}

    \item \textit{Enhanced User Accessibility and Engagement:}  
    LCMs offer personalized and context-aware recommendations, making digital systems more inclusive and engaging.  
    \begin{itemize}[leftmargin=*]
        \item {Contextual Feedback:} E-learning platforms can offer tailored feedback to users based on their progress and learning goals by using LCMs.
        \item {Language Localization:} Practitioners can translate and localize educational or medical content into low-resource languages with LCMs, making critical information accessible to more users.
        \item {Inclusive Communication:} By catering to diverse linguistic and cultural needs, LCM-powered systems ensure that users from underserved communities receive accurate and relevant information.
    \end{itemize}
    \vspace{0.05in}

    \item \textit{Regulatory Compliance and Legal Efficiency:}  
    LCMs enable practitioners to enhance legal workflows by comparing policies, summarizing legal documents, and identifying regulatory clauses to ensure compliance \cite{Ryan_lcm}.
    \begin{itemize}[leftmargin=*]
        \item \textbf{Policy Analysis and Comparison:} Practitioners can utilize LCMs to compare legal policies across jurisdictions, identifying key differences and commonalities.
        \item \textbf{Automated Regulatory Monitoring:} By leveraging LCMs, practitioners can track changes in regulations and provide timely updates to compliance teams, ensuring policies are adjusted accordingly.
        \item \textbf{Legal Document Summarization:} Legal professionals can save time by accessing summaries of lengthy contracts and court rulings generated by LCMs \cite{Ryan_lcm}.
    \end{itemize}
    \vspace{0.05in}

    \item \textit{Medical Information Processing:}  
    LCMs empower practitioners to process complex medical documents, supporting efficient summarization, comparison, and translation tasks.
    \begin{itemize}[leftmargin=*]
        \item \textbf{Patient Record Summarization:} Practitioners can create concise overviews of patient histories using LCMs, aiding quick clinical decision-making.
        \item \textbf{Clinical Trial Analysis:} Research institutions can employ LCMs to compare clinical trial outcomes and identify emerging trends.
        \item \textbf{Cross-Lingual Medical Support:} LCMs facilitate the translation of medical reports and discharge instructions, enhancing care coordination for non-native-speaking patients.
    \end{itemize}
    \vspace{0.05in}

    \item \textit{Creative and Collaborative Content Generation:}  
    LCMs assist content creators in generating structured story expansions and maintaining thematic consistency in collaborative writing projects.
    \begin{itemize}[leftmargin=*]
        \item \textbf{Story Expansion:} Media production teams can use LCMs to develop detailed storylines and scripts based on initial outlines.
        \item \textbf{Collaborative Reports:} News agencies can co-author articles in real time with the help of LCMs, ensuring timely and accurate event reporting.
        \item \textbf{Conceptual Feedback:} Educational institutions can provide conceptual feedback on research papers and student assignments using LCMs.
    \end{itemize}
    \vspace{0.05in}

    \item \textit{Efficient Knowledge Management:}  
    LCMs improve knowledge retrieval and management by creating semantic links between documents, making information more accessible.
    \begin{itemize}[leftmargin=*]
        \item \textbf{Semantic Search:} Practitioners can use LCM-powered search engines to find relevant documents even if they are phrased differently across files.
        \item \textbf{Document Linking and Annotation:} LCMs allow for the linking of related reports and annotation of key sections, providing additional context.
        \item \textbf{Centralized Knowledge Hubs:} Organizations can build comprehensive knowledge bases with LCMs that map concept-level relationships between documents.
    \end{itemize}
    \vspace{0.05in}

    \item \textit{Improved Customer Interaction:}  
    Practitioners can enhance customer service by utilizing LCMs to support multilingual and multimodal chatbots, offering more personalized interactions.
    \begin{itemize}[leftmargin=*]
        \item \textbf{Multilingual Chatbots:} Practitioners can deploy LCM-powered chatbots that respond to user queries in multiple languages, ensuring consistent support across different regions.
        \item \textbf{Speech-to-Text Support:} By integrating LCMs, practitioners can enable users to submit voice queries for real-time assistance, thereby improving accessibility for visually impaired customers.
        \item \textbf{Context-Aware Responses:} Leveraging LCMs' understanding of query context, practitioners can provide more relevant and helpful responses, enhancing overall customer satisfaction.
    \end{itemize}
    \vspace{0.05in}

    \item \textit{E-Learning and Training:}  
    Practitioners can utilize LCMs to provide personalized training modules and real-time feedback, thereby improving learning outcomes.
    \begin{itemize}[leftmargin=*]
        \item \textbf{Adaptive Learning Modules:} Practitioners can generate lesson plans and study notes tailored to individual learners' progress using LCMs, ensuring a customized educational experience.
        \item \textbf{Interactive Feedback:} Educational platforms can employ LCMs to offer detailed sentence-level feedback on written assignments, aiding in student development and performance.
        \item \textbf{Knowledge Gap Identification:} By analyzing learner data, practitioners can use LCMs to identify areas where learners need improvement and suggest targeted exercises to address these gaps.
    \end{itemize}
    \vspace{0.05in}

    \item \textit{Decision Support Systems:}  
    Practitioners can leverage LCMs to assist in decision-making by summarizing reports, generating action plans, and providing hierarchical overviews of complex data \cite{Zenspire7_lcm}.
    \begin{itemize}[leftmargin=*]
        \item \textbf{Policy Recommendations:} Public sector agencies can rely on LCMs to draft policy briefs and identify potential impacts of proposed regulations, aiding in informed policy-making.
        \item \textbf{Financial Insights:} Financial practitioners can use LCMs to analyze financial reports and market trends, suggesting investment strategies based on comprehensive data analysis.
        \item \textbf{Strategic Planning:} Organizations can employ LCM-generated summaries of market research to inform their decision-making processes and future initiatives, ensuring strategic alignment with market dynamics.
    \end{itemize}

\end{enumerate}

\begin{custombox}
LCMs enable practitioners to enhance workflows through automation, cross-lingual support, and personalized user engagement. They allow for improved regulatory compliance, medical analysis, and knowledge management by utilizing semantic search and generating accurate summaries. Additionally, practitioners can strengthen customer interactions and e-learning initiatives through context-aware, multimodal support provided by LCMs. In essence, LCMs empower practitioners to optimize processes, enhance decision-making, and deliver more adaptive, user-centric solutions across various industries.
\end{custombox}

\section{Potential Limitations of LCM} \label{Section 5}
LCM approach introduces a paradigm shift by emphasizing conceptual reasoning across multilingual and multimodal contexts. However, as reported in \cite{the2024large}, several inherent limitations impact its effectiveness, necessitating further exploration and enhancements. Below is a detailed discussion of these limitations, organized into key areas.

\begin{enumerate}[leftmargin=*]
\item \textit{Embedding Space Design:}
The performance of LCMs relies on the quality of the embedding space. The SONAR embedding space, chosen for its strong multilingual capabilities, was trained on bitext machine translation data with short sentences, creating a distribution mismatch with real-world corpora. This leads to challenges in predicting sequences of loosely related sentences and handling content such as links, references, or numerical data. Additionally, using a frozen encoder ensures stable representations but may be suboptimal for tasks requiring end-to-end adaptability. Joint training could enhance performance but at the cost of increased computational requirements and potential modality competition.

\vspace{0.1in}

    \item \textit{Concept Granularity:}
LCMs interpret concepts at the sentence level, which can limit their ability to represent longer sentences with multiple ideas. The combinatorial complexity of possible next sentences grows exponentially with sentence length, making it difficult to assign accurate probabilities for continuations. Additionally, the sparsity of unique sentences in large corpora impacts generalization. While sentence splitting or mapping sentences into smaller conceptual units could improve generalization, creating universally applicable conceptual units across languages and modalities remains a challenge.

\vspace{0.1in}

    \item \textit{Continuous vs. Discrete Representations:}
While diffusion-based modeling excels in continuous data like images and speech, it struggles with text due to its discrete structure. Sentence embeddings, though continuous vectors, correspond to discrete language constructs, making generation difficult. Additionally, diffusion models lack the contrastive loss mechanisms that improve performance in tasks such as code generation and multiple-choice questions. Although quantization (as in Quant-LCM) aims to address this, the current SONAR embedding space is not optimized for efficient quantization, leading to an explosion of possible combinations and limiting performance.

\vspace{0.1in}

    \item \textit{Generalization Across Languages and Modalities:}
Generalizing across languages and modalities requires constructing conceptual units that are shared across diverse inputs. However, building a comprehensive multilingual and multimodal dataset is resource-intensive and challenging. Effective generalization must strike a balance between preserving key details (such as named entities) and enabling abstraction for reasoning. LCMs need exposure to more diverse datasets to enhance their ability to transfer knowledge across languages and modalities.
\end{enumerate}

\section{Conclusion}  \label{Section 6}

This study investigated the emerging paradigm of LCMs, distinguishing them from traditional token-based LLMs. Unlike conventional models that process text one token at a time, LCMs operate at the conceptual level, treating entire sentences or ideas as unified semantic units. This approach enhances interpretability, enables more effective reasoning over extended contexts, and offers adaptability across various languages and modalities. By synthesizing insights from grey literature, we identified the defining characteristics, key use cases, and implications of LCMs for both researchers and practitioners. Our findings illustrate that a distinguishing strength of LCMs is their ability to function within a language and modality-agnostic conceptual space, facilitating efficient long-context reasoning and cross-domain applications. Moreover, our study demonstrates the significant versatility of LCMs in diverse domains such as cybersecurity, healthcare, and education, where they improve decision-making, enhance resource efficiency, and support innovative multimodal solutions. 

LCMs face several challenges, including the need for robust embedding spaces, precise concept granularity, and managing trade-offs between continuous and discrete data representations. However, addressing these challenges presents opportunities to develop refined embeddings, enhance quantization strategies, and create cross-domain frameworks that leverage LCMs for more interpretable and context-sensitive artificial intelligence. As a result, LCMs are poised to transform the next generation of AI applications. Future research will likely focus on overcoming current limitations and refining architectural designs to harness concept-driven modelling fully. By advancing LCM technology, researchers and practitioners can promote more inclusive communication, accelerate cross-disciplinary collaboration, and drive AI toward greater interpretability, efficiency, and contextual intelligence.

\bibliographystyle{IEEEtran}
\bibliography{ref}

\begin{thebibliography}{10}
\providecommand{\url}[1]{#1}
\csname url@samestyle\endcsname
\providecommand{\newblock}{\relax}
\providecommand{\bibinfo}[2]{#2}
\providecommand{\BIBentrySTDinterwordspacing}{\spaceskip=0pt\relax}
\providecommand{\BIBentryALTinterwordstretchfactor}{4}
\providecommand{\BIBentryALTinterwordspacing}{\spaceskip=\fontdimen2\font plus
\BIBentryALTinterwordstretchfactor\fontdimen3\font minus \fontdimen4\font\relax}
\providecommand{\BIBforeignlanguage}[2]{{%
\expandafter\ifx\csname l@#1\endcsname\relax
\typeout{** WARNING: IEEEtran.bst: No hyphenation pattern has been}%
\typeout{** loaded for the language `#1'. Using the pattern for}%
\typeout{** the default language instead.}%
\else
\language=\csname l@#1\endcsname
\fi
#2}}
\providecommand{\BIBdecl}{\relax}
\BIBdecl

\bibitem{zheng2025towards}
Z.~Zheng, K.~Ning, Q.~Zhong, J.~Chen, W.~Chen, L.~Guo, W.~Wang, and Y.~Wang, ``Towards an understanding of large language models in software engineering tasks,'' \emph{Empirical Software Engineering}, vol.~30, no.~2, p.~50, 2025.

\bibitem{chang2024survey}
Y.~Chang, X.~Wang, J.~Wang, Y.~Wu, L.~Yang, K.~Zhu, H.~Chen, X.~Yi, C.~Wang, Y.~Wang \emph{et~al.}, ``A survey on evaluation of large language models,'' \emph{ACM Transactions on Intelligent Systems and Technology}, vol.~15, no.~3, pp. 1--45, 2024.

\bibitem{haque2022think}
M.~U. Haque, I.~Dharmadasa, Z.~T. Sworna, R.~N. Rajapakse, and H.~Ahmad, ``" i think this is the most disruptive technology": Exploring sentiments of chatgpt early adopters using twitter data,'' \emph{arXiv preprint arXiv:2212.05856}, 2022.

\bibitem{Mehul_LCM}
{Mehul Gupta}, ``Meta large concept models (lcm): End of llms?'' \url{https://medium.com/data-science-in-your-pocket/meta-large-concept-models-lcm-end-of-llms-68cb0c5cd5cf}, accessed: January 6, 2025.

\bibitem{huang2023advancing}
Y.~Huang, J.~Xu, J.~Lai, Z.~Jiang, T.~Chen, Z.~Li, Y.~Yao, X.~Ma, L.~Yang, H.~Chen \emph{et~al.}, ``Advancing transformer architecture in long-context large language models: A comprehensive survey,'' \emph{arXiv preprint arXiv:2311.12351}, 2023.

\bibitem{borgeaud2022improving}
S.~Borgeaud, A.~Mensch, J.~Hoffmann, T.~Cai, E.~Rutherford, K.~Millican, G.~B. Van Den~Driessche, J.-B. Lespiau, B.~Damoc, A.~Clark \emph{et~al.}, ``Improving language models by retrieving from trillions of tokens,'' in \emph{International conference on machine learning}.\hskip 1em plus 0.5em minus 0.4em\relax PMLR, 2022, pp. 2206--2240.

\bibitem{chopra2024chatnvd}
S.~Chopra, H.~Ahmad, D.~Goel, and C.~Szabo, ``Chatnvd: Advancing cybersecurity vulnerability assessment with large language models,'' \emph{arXiv preprint arXiv:2412.04756}, 2024.

\bibitem{gereti2024token}
M.~Gereti, A.~Robinson, S.~Williams, C.~Anderson, and D.~Walker, ``Token-based prompt manipulation for automated large language model evaluation,'' \emph{Authorea Preprints}, 2024.

\bibitem{kim2025palm}
S.~Kim, D.~Huang, Y.~Xian, O.~Hilliges, L.~Van~Gool, and X.~Wang, ``Palm: Predicting actions through language models,'' in \emph{European Conference on Computer Vision}.\hskip 1em plus 0.5em minus 0.4em\relax Springer, 2025, pp. 140--158.

\bibitem{Lance_lcm}
{Lance Eliot}, ``Ai is breaking free of token-based llms by upping the ante to large concept models that devour sentences and adore concepts,'' \url{https://www.forbes.com/sites/lanceeliot/2025/01/06/ai-is-breaking-free-of-token-based-llms-by-upping-the-ante-to-large-concept-models-that-devour-sentences-and-adore-concepts}, accessed: January 06, 2025.

\bibitem{LLM_Limit}
{Jacob Grow}, ``The strengths and limitations of large language models in reasoning, planning, and code integration,'' \url{https://medium.com/@Gbgrow/the-strengths-and-limitations-of-large-language-models-in-reasoning-planning-and-code-41b7a190240c}, accessed: January 5, 2025.

\bibitem{cuskley2024limitations}
C.~Cuskley, R.~Woods, and M.~Flaherty, ``The limitations of large language models for understanding human language and cognition,'' \emph{Open Mind}, vol.~8, pp. 1058--1083, 2024.

\bibitem{LLM_Limit22}
{Bijit Ghosh}, ``Data design for fine-tuning llms with long context windows,'' \url{https://medium.com/@bijit211987/data-design-for-fine-tuning-llms-with-long-context-windows-4f560afbd1d0}, accessed: January 5, 2025.

\bibitem{LLM_Limit333}
{DrKilngon}, ``Limitations of llm models,'' \url{https://medium.com/@dbiswajitdatta/limitations-of-llm-models-03cc3d6645b6}, accessed: January 5, 2025.

\bibitem{fu2024moa}
T.~Fu, H.~Huang, X.~Ning, G.~Zhang, B.~Chen, T.~Wu, H.~Wang, Z.~Huang, S.~Li, S.~Yan \emph{et~al.}, ``Moa: Mixture of sparse attention for automatic large language model compression,'' \emph{arXiv preprint arXiv:2406.14909}, 2024.

\bibitem{petrick2022locality}
F.~Petrick, J.~Rosendahl, C.~Herold, and H.~Ney, ``Locality-sensitive hashing for long context neural machine translation,'' in \emph{Proceedings of the 19th International Conference on Spoken Language Translation (IWSLT 2022)}, 2022, pp. 32--42.

\bibitem{the2024large}
L.~The, L.~Barrault, P.-A. Duquenne, M.~Elbayad, A.~Kozhevnikov, B.~Alastruey, P.~Andrews, M.~Coria, G.~Couairon, M.~R. Costa-juss{\`a} \emph{et~al.}, ``Large concept models: Language modeling in a sentence representation space,'' \emph{arXiv preprint arXiv:2412.08821}, 2024.

\bibitem{video_lcm}
{AI Papers Academy}, ``Large concept models (lcms) by meta: The era of ai after llms?'' \url{https://www.youtube.com/watch?v=TwLiNTYvpPo}, accessed: January 4, 2025.

\bibitem{LLMsKnow}
{Javaid Nabi}, ``All you need to know about llm text generations,'' \url{https://medium.com/@javaid.nabi/all-you-need-to-know-about-llm-text-generation-03b138e0ed19}, accessed: January 5, 2025.

\bibitem{Richard_lcm}
{Richard Aragon}, ``Experimenting with lcm models (meta's alternative to llm models),'' \url{https://www.youtube.com/watch?v=2ZLd0uZvwbU}, accessed: January 03, 2025.

\bibitem{LCMsss}
{Vimal Prakash}, ``Large concept models (lcm),'' \url{https://www.linkedin.com/pulse/large-concept-models-lcm-vimal-prakash-pjgac}, accessed: January 3, 2025.

\bibitem{lcm2024}
{Zen}, ``Large concept models (lcms),'' \url{https://medium.com/@ThisIsMeIn360VR/large-concept-models-lcms-d59b86531ef6}, accessed: January 5, 2025.

\bibitem{pruseth_lcm_agi}
{Debabrata Pruseth}, ``Lcms (large concept models): The path to agi, the future of ai thinking,'' \url{https://debabratapruseth.com/lcms-large-concept-models-the-path-to-agi-the-future-of-ai-thinking/}, accessed: January 5, 2025.

\bibitem{ashish_lcm}
{Ashish Bamaniah}, ``Meta’s large concept models (lcms) are here to challenge and redefine llms,'' \url{https://levelup.gitconnected.com/metas-large-concept-models-lcms-are-here-to-challenge-and-redefine-llms-7f9778f88a87}, accessed: January 5, 2025.

\bibitem{Asif_lcm}
{Asif Razzaq}, ``Meta ai proposes large concept models (lcms): A semantic leap beyond token-based language modeling,'' \url{https://www.marktechpost.com/2024/12/15/meta-ai-proposes-large-concept-models-lcms-a-semantic-leap-beyond-token-based-language-modeling}, accessed: January 02, 2025.

\bibitem{annirudhha_lcm}
{Aniruddha Shrikhande}, ``A deep dive into large concept models (lcms),'' \url{https://adasci.org/a-deep-dive-into-large-concept-models-lcms}, accessed: January 6, 2025.

\bibitem{Simplify_LCM}
{Simplify AI}, ``Meta's large concept models (lcms): The future of ai innovation!'' \url{https://www.youtube.com/watch?v=PhXusDEcXQk}, accessed: January 7, 2025.

\bibitem{Jiaxin_lcm}
{Jiaxin Bai}, ``What is the concepts in large concept model and why its useful,'' \url{https://bjx.fun/p/what-is-the-concepts-in-large-concept-model-and-why-its-useful}, accessed: January 03, 2025.

\bibitem{Aniket_LCM}
{Aniket Hingane}, ``Practical advancements in ai: How large concept models are redefining the landscape of llms,'' \url{https://medium.com/@learn-simplified/practical-advancements-in-ai-how-large-concept-models-are-redefining-the-landscape-of-llms-b0220296458b}, accessed: January 2, 2025.

\bibitem{Siddhant_lcm}
{Siddhant Rai and Vizuara AI}, ``Large concept models : Language modeling in a sentence representation space,'' \url{https://vizuara.substack.com/p/large-concept-models-language-modeling?}, accessed: January 01, 2025.

\bibitem{Ganesh_LCM}
{Ganesh Raju}, ``The next evolution of ai: Trading tokens for concepts - large concept models,'' \url{https://www.linkedin.com/pulse/next-evolution-ai-trading-tokens-concepts-large-concept-ganesh-raju-cdgwc}, accessed: January 02, 2025.

\bibitem{Fahd_lcm}
{Fahd Mirza}, ``Meta introduces large concept models,'' \url{https://www.youtube.com/watch?v=O-uHwOTkubc}, accessed: January 4, 2025.

\bibitem{aii_LCM}
{ai explained}, ``Future of llms? meta's new lcms explained (large concept models),'' \url{https://www.youtube.com/watch?v=Si04vFKjL1s}, accessed: January 2, 2025.

\bibitem{Wes_lcm}
{Wes Roth}, ``Meta's stunning new llm architecture is a game-changer!'' \url{https://www.youtube.com/watch?v=jvdt4jRKxOg}, accessed: January 01, 2025.

\bibitem{Leadership_LCM}
{Leadership in AI}, ``Meta's stunning lcm large concept models for artificial intelligence | they are thinking now!'' \url{https://www.youtube.com/watch?v=u_Z3HCw8ApQ}, accessed: January 8, 2025.

\bibitem{Anaam_lcm}
{Anaam Rasool}, ``Large concept models: The future of ai is here!'' \url{https://www.youtube.com/watch?v=Y63fH1sO_tg}, accessed: January 06, 2025.

\bibitem{BazAI_LCM}
{BazAI}, ``what is lcm ? large concept models : Beyond token-level processing,'' \url{https://www.youtube.com/watch?v=RcN_s819_Ns}, accessed: January 8, 2025.

\bibitem{TheAIGRID_LCM}
{TheAIGRID}, ``Experts are stunned! meta's new llm architecture is a game-changer!'' \url{https://www.youtube.com/watch?v=1Z1vKdrmpj4}, accessed: January 2, 2025.

\bibitem{TheAILabsCanada_LCM}
{TheAILabsCanada}, ``Meta introduces large concept models (lcm) - the end of large language models (llms)?'' \url{https://www.youtube.com/watch?v=s4JsTlBDwNo}, accessed: January 3, 2025.

\bibitem{Discover_LCM}
{Discover AI}, ``Lcm: The ultimate evolution of ai? large concept models,'' \url{https://www.youtube.com/watch?v=y1MG0BCf3UU}, accessed: January 1, 2025.

\bibitem{Next_LCM}
{NextSprints}, ``Large concept models, training inference and applications,'' \url{https://www.youtube.com/watch?v=HXrpVXOYuV0}, accessed: January 8, 2025.

\bibitem{Quantum_LCM}
{Quantum Circuit}, ``Meta’s latest llm architecture shocks experts,'' \url{https://www.youtube.com/watch?v=clkKDS-hPN4}, accessed: January 7, 2025.

\bibitem{Teendifferent_LCM}
{Teendifferent}, ``What are large concept models? the ai innovation you need to know about!!'' \url{https://medium.com/predict/what-are-large-concept-models-the-meta-ai-innovation-you-need-to-know-about-2375a618fed1}, accessed: January 8, 2025.

\bibitem{DataScience_lcm}
{Data Science in your pocket}, ``Meta large concept models (lcms),'' \url{https://www.youtube.com/watch?v=GY-UGAsRF2g}, accessed: January 06, 2025.

\bibitem{Swaroop_lcm}
{Swaroop Piduguralla}, ``Rethinking language models: The emergence of large concept models,'' \url{https://medium.com/@tejaswaroop2310/rethinking-language-models-the-emergence-of-large-concept-models-76746acfea13}, accessed: January 03, 2025.

\bibitem{Albert_LCM}
{Albert Harmon}, ``Exploring large concept models: The next evolution in ai,'' \url{https://galaxy.ai/youtube-summarizer/exploring-large-concept-models-the-next-evolution-in-ai-y1MG0BCf3UU}, accessed: January 01, 2025.

\bibitem{Bhavik_lcm}
{Bhavik Jikadara}, ``Meta's large concept models (lcms) redefine nlp,'' \url{https://medium.com/ai-agent-insider/metas-large-concept-models-lcms-redefine-nlp-32167e7ddb6c}, accessed: January 5, 2025.

\bibitem{Ajith_lcm}
{Ajith Prabhakar}, ``Large concept model (lcm): Redefining language understanding with multilingual and modality-agnostic ai,'' \url{https://ajithp.com/2025/01/05/large-concept-model-lcm-redefining-language-understanding-with-multilingual-and-modality-agnostic-ai}, accessed: January 5, 2025.

\bibitem{Keven_LCM}
{KevenBazile}, ``Why meta's lcm models are the secret to unlocking ai's true potential,'' \url{https://www.youtube.com/watch?v=LQtn0tFct0s}, accessed: January 7, 2025.

\bibitem{AIPapers_lcm}
{AI Papers Podcast Daily}, ``Large concept models: Language modeling in a sentence representation space,'' \url{https://www.youtube.com/watch?v=t3LSYprkDLo}, accessed: January 1, 2025.

\bibitem{Zenspire7_lcm}
{Zenspire7}, ``Meta's large concept models: A new paradigm in ai language understanding,'' \url{https://www.youtube.com/watch?v=kwYatIgL0Ek}, accessed: January 03, 2025.

\bibitem{bjx_fun_lcm}
{BJX Fun}, ``What is the concept in large concept models and why it's useful?'' \url{https://bjx.fun/p/what-is-the-concepts-in-large-concept-model-and-why-its-useful}, accessed: January 5, 2025.

\bibitem{Income_lcm}
{Income stream surfers}, ``Are large concept models the future of ai? breaking down meta’s lcm,'' \url{https://www.youtube.com/watch?v=ZTldUaf10IQ}, accessed: January 02, 2025.

\bibitem{Anthony_LCM}
{Anthony Repetto}, ``Beyond meta: Large concept models will win,'' \url{https://medium.com/predict/beyond-meta-large-concept-models-will-win-05728a8f4074}, accessed: January 01, 2025.

\bibitem{Synced_lcm}
{Synced}, ``From token to conceptual: Meta introduces large concept models in multilingual ai,'' \url{https://medium.com/syncedreview/from-token-to-conceptual-the-rise-of-metas-large-concept-models-in-multilingual-ai-b32acbfeb792}, accessed: January 4, 2025.

\bibitem{Absaar_lcm}
{Absaar}, ``Meta's shift from llm (large language models) to lcm (large concept models),'' \url{https://www.youtube.com/watch?v=ZcVMajfdilI}, accessed: January 05, 2025.

\bibitem{szymonslowik_lcm}
{Szymon Slowik}, ``Large concept models (lcms) explained,'' \url{https://www.szymonslowik.com/large-concept-models-lcms-explained}, accessed: January 5, 2025.

\bibitem{viliotti_lcm_semantic_reasoning}
{Andrea Viliotti}, ``Large concept model (lcm): A new paradigm for large-scale semantic reasoning in ai,'' \url{https://www.andreaviliotti.it/post/large-concept-model-lcm-a-new-paradigm-for-large-scale-semantic-reasoning-in-ai/}, accessed: January 5, 2025.

\bibitem{AIbase_LCM}
{AIbase}, ``Meta launches 'large concept models' (lcms)! breaking through llm limitations and leading a new direction in ai language understanding,'' \url{https://www.aibase.com/news/13985}, accessed: January 5, 2025.

\bibitem{Py_LCM}
{Py Man}, ``Meta giving llms human thinking! (lcm explained),'' \url{https://www.youtube.com/watch?v=caOffcqG_GQ}, accessed: January 5, 2025.

\bibitem{Daniele_lcm}
{Daniele Moltisanti}, ``Large concept models: Meta’s next frontier in ai,'' \url{https://staituned.com/learn/expert/large-concept-model-meta}, accessed: January 03, 2025.

\bibitem{TOPAI_LCM}
{TOP AI}, ``The big leap in ai: Meta's large concept models,'' \url{https://www.youtube.com/watch?v=HQ91uiOaAEM}, accessed: January 7, 2025.

\bibitem{Arman_LCM}
{Arman Kamran}, ``Large concept models (lcms): A conceptual leap in generative ai,'' \url{https://medium.com/@armankamran/large-concept-models-lcms-a-conceptual-leap-in-generative-ai-29a9a7256643}, accessed: January 8, 2025.

\bibitem{Ryan_lcm}
{Ryan McDonough}, ``Rethinking ai in legal tech: The role of large concept models,'' \url{https://www.ryanmcdonough.co.uk/rethinking-ai-in-legal-tech-the-role-of-large-concept-models}, accessed: January 03, 2025.

\bibitem{abdulsatar2024towards}
M.~Abdulsatar, H.~Ahmad, D.~Goel, and F.~Ullah, ``Towards deep learning enabled cybersecurity risk assessment for microservice architectures,'' \emph{arXiv preprint arXiv:2403.15169}, 2024.

\bibitem{ahmad2024survey}
H.~Ahmad, F.~Ullah, and R.~Jafri, ``A survey on immersive cyber situational awareness systems,'' \emph{arXiv preprint arXiv:2408.07456}, 2024.

\bibitem{jayalath2024microservice}
R.~K. Jayalath, H.~Ahmad, D.~Goel, M.~S. Syed, and F.~Ullah, ``Microservice vulnerability analysis: A literature review with empirical insights,'' \emph{IEEE Access}, 2024.

\bibitem{goel2024machine}
D.~Goel, H.~Ahmad, A.~K. Jain, and N.~K. Goel, ``Machine learning driven smishing detection framework for mobile security,'' \emph{arXiv preprint arXiv:2412.09641}, 2024.

\bibitem{ahmad2023review}
H.~Ahmad, I.~Dharmadasa, F.~Ullah, and M.~A. Babar, ``A review on c3i systems’ security: Vulnerabilities, attacks, and countermeasures,'' \emph{ACM Computing Surveys}, vol.~55, no.~9, pp. 1--38, 2023.

\bibitem{syed2020cybersecurity}
R.~Syed, ``Cybersecurity vulnerability management: A conceptual ontology and cyber intelligence alert system,'' \emph{Information \& Management}, vol.~57, no.~6, p. 103334, 2020.

\bibitem{ahmad2024towards}
H.~Ahmad, C.~Treude, M.~Wagner, and C.~Szabo, ``Towards resource-efficient reactive and proactive auto-scaling for microservice architectures,'' \emph{Available at SSRN 4918202}, 2024.

\bibitem{Ahmad_2024}
\BIBentryALTinterwordspacing
------, ``Smart hpa: A resource-efficient horizontal pod auto-scaler for microservice architectures,'' in \emph{2024 IEEE 21st International Conference on Software Architecture (ICSA)}.\hskip 1em plus 0.5em minus 0.4em\relax IEEE, Jun. 2024, p. 46–57. [Online]. Available: \url{http://dx.doi.org/10.1109/ICSA59870.2024.00013}
\BIBentrySTDinterwordspacing

\bibitem{sox2024medical}
H.~C. Sox, M.~C. Higgins, D.~K. Owens, and G.~S. Schmidler, \emph{Medical decision making}.\hskip 1em plus 0.5em minus 0.4em\relax John Wiley \& Sons, 2024.

\bibitem{abbas2024robust}
F.~Abbas and H.~Ahmad, ``Robust partial least squares using low rank and sparse decomposition,'' \emph{arXiv preprint arXiv:2407.06936}, 2024.

\end{thebibliography}
\end{document}